\documentclass[11pt]{article}

\usepackage[preprint]{acl}
\usepackage{times}
\usepackage{latexsym}
\usepackage[T1]{fontenc}
\usepackage[utf8]{inputenc}
\usepackage{microtype}
\usepackage{graphicx}
\usepackage{booktabs}
\usepackage{amsmath}
\usepackage{amsfonts}
\usepackage{pifont}
\usepackage{enumitem}
\usepackage{listings}
\usepackage{multirow}
\usepackage{xcolor}
\usepackage{array}
\usepackage{rotating}
\usepackage{xspace}
\usepackage{nicefrac}

\newcommand{\cmark}{\ding{51}}

\newcommand{\Ind}{\mathbf{1}}

\title{Evaluating Deep Research Agents on Expert Consulting Work:\\
A Benchmark with Verifiers, Rubrics, and Cognitive Traps}

\author{
  Tanmay Asthana \quad Aman Saksena \quad Divyansh Sahu \\
  Deccan AI Research \\
  \texttt{tanmay.asthana@deccan.ai}
}

\begin{document}

\maketitle

\begin{abstract}
Frontier deep research agents (DRAs) are being deployed in enterprise
workflows faster than they are being evaluated. Existing benchmarks
measure factual recall, single-hop QA, or generic agentic skill, and
miss the multi-document, decision-grade deliverables DRAs are asked to
produce. We introduce a benchmark of 70 SME-authored management
consulting prompts, each embedding cognitive traps that penalize
surface-pattern reasoning. Three frontier agents, namely Claude Opus~4.6,
OpenAI o3-deep-research and Gemini~3.1~Pro deep-research, are scored on
two complementary layers: deterministic binary verifiers (mean 14.9
per task) and a five-criterion 0--3 SME rubric (Data Integrity,
Analytical Rigor, Relevance \& Focus, Execution Precision, Format \&
Deliverability), combined into a Verifier-Rubric Score (VRS, 0--100).

Acceptance under a joint threshold (rubric mean $\geq 2.5$ and
verifier pass rate $\geq 80\%$) is uniformly low: o3 15.7\%, Claude
12.9\%, Gemini 12.9\%. Pairwise differences are statistically
indistinguishable. On the continuous VRS, o3 leads
(61.4~[CI: 55.2,\,67.5]), followed by Gemini (52.6) and Claude (38.5); the
o3--Claude gap ($\Delta{=}22.9$, $p{<}0.001$) survives Bonferroni
correction. No agent averages above the rubric's ``adequate''
threshold of 2.0; no agent's mean verifier pass rate reaches the 80\%
acceptance floor. Each agent fails distinctively: Claude leads on data
fabrication and file-access failures; o3 propagates cascading
computation errors; Gemini oscillates between the highest
perfect-verifier rate and the most catastrophic collapses. The
benchmark, evaluation code, and full prompt corpus are publicly
released.
\end{abstract}

\section{Introduction}
\label{sec:intro}

The companies that sell deep research agents have moved faster than the people evaluating them. DRAs are already being wired into enterprise pipelines where the answers feed multi-million-dollar decisions. Most of the benchmarks used to vet these systems were not built for that kind of use. The dominant ones measure factual recall (MMLU~\cite{hendrycks2021mmlu}), single-hop question answering (TriviaQA~\cite{joshi2017triviaqa}), web navigation (WebArena~\cite{zhou2023webarena}), or generic agentic skill (GAIA~\cite{mialon2023gaia}, AgentBench~\cite{liu2023agentbench}). The recent wave of professional-domain benchmarks like FinanceBench~\cite{islam2023financebench} for finance, LegalBench~\cite{guha2023legalbench} for law, MedQA~\cite{jin2021medqa} for medicine, is a step in the right direction. However, these still frame evaluation as question-answering rather than the production of decision-grade structured deliverables. The methodological literature on agentic evaluation~\cite{xi2023agentsurvey, gu2024judgesurvey, liang2023helm, srivastava2022bigbench} has flagged exactly this gap.

The cost of the gap is concrete. One task in our corpus involving an inbound-freight cost program for a packaging firm would commit the company to roughly \texteuro4.5~billion of capital expenditure on a defective basis, if a single Year-5 segment revenue figure is miscalculated. Modern deep research agents have measurable rates of exactly this kind of miscalculation, and well-documented tendencies to confabulate when the source material is silent~\cite{ji2023hallucination, huang2023halusurvey, kadavath2022calibration}.

Our benchmark comprises 70 tasks tested over 3 agents: Claude Opus 4.6 with web search, OpenAI
o3-deep-research, and Google Gemini 3.1 Pro deep-research. It is built around three design choices that distinguish it from prior research-agent benchmarks:

\begin{itemize}
    \item The scoring is two-layered: every response is checked by a suite of binary task-specific verifiers and then independently scored by a subject-matter expert on a five-criterion 0-3 rubric (Data Integrity, Analytical Rigor, Relevance \& Focus, Execution Precision, Format \& Deliverability), with the two layers combined into a Verifier-Rubric Score (VRS) on 0-100. The dual layer exposes agent-distinct failures that single-metric benchmarks systematically miss (Section~\ref{sec:results:failures}).
    \item The prompt corpus is organized not by topic but by cognitive capability: five classes targeting Constrained Research Prompts (CRP), Relevance Compression Prompts (RCP), Structural Compliance Prompts (SCP), Latent Decomposition Prompts (LDP), and Failure-Sensitive Prompts (FSP), each isolating a kind of reasoning we wanted to test independently (Sections~\ref{sec:results:per_type}--\ref{sec:results:effect_sizes}).
    \item Many of the prompts deliberately embed cognitive traps in the form of realistic human-style errors in the source documents (inconsistent units, footnote-body contradictions, non-standard date formats) and deterministic precision failures (averaging where weighting is required, defaulting to plausible numbers in the absence of an explicit value). These keep the benchmark hard for shallow heuristics; a benchmark with only clean inputs would systematically understate the difficulty of professional research work.
\end{itemize}

The framework of our benchmark captures distinct failure modes besides just providing a comparison of success rates. At $n{=}70$, o3 leads on continuous VRS (61.4), reasoning mean (1.89), and fewest criterion-zeros (35), yet ties the other agents on the binary ACCEPT threshold at $\sim$13--16\%. Claude has the highest auto-reject rate (41.4\%) driven by 22 Data Integrity zeros and 62 total criterion-zeros. Gemini accumulates the most criterion-zeros (74) but wins the most per-prompt VRS argmax matches (26.3 of 70) and leads on two prompt types. The continuous and binary views diverge: continuous VRS distinguishes agents, but the stringent ACCEPT threshold ($\bar{r} \geq 2.5$ and $V \geq 80\%$) compresses them into statistical indistinguishability. These per-agent signatures, the prompt-class-conditional effect sizes, and the criterion correlation structure together support this benchmark as a discriminating evaluation framework for frontier deep research agents.

Section~\ref{sec:related} situates the benchmark against related work. Section~\ref{sec:benchmark} describes the prompt taxonomy, cognitive-trap design, dual-layer scoring framework, and multi-agent infrastructure. Section~\ref{sec:results} reports headline results, per-type performance, failure signatures, and cross-benchmark validation. Section~\ref{sec:limitations} discusses caveats; Section~\ref{sec:conclusion} concludes. Extended diagnostics, worked examples, and the full QC protocol are in the appendices. 

\textbf{Reproducibility:} All evaluation code can be found here. 

Code: \url{https://anonymous.4open.science/status/dra-response-gen-4288}. 

Dataset (70-prompt corpus and grading workbook) will be released publicly on paper publication.

\section{Related Work}
\label{sec:related}

\textbf{Research agent and domain-specific benchmarks.}~General-purpose agentic benchmarks (GAIA~\cite{mialon2023gaia}, BrowseComp~\cite{browsecomp2025}, WebArena~\cite{zhou2023webarena}, AgentBench~\cite{liu2023agentbench}, ToolBench~\cite{qin2023toolllm}, ToolEval~\cite{patil2023gorilla}) target tool use, browsing, and multi-step reasoning but not domain-expert research. Deep-research benchmarks (DeepResearch Bench~\cite{du2025deepresearchbench}, ResearcherBench~\cite{xu2025researcherbench}, SciAgent~\cite{ma2024sciagent}, SciBench~\cite{wang2024scibench}) target scientific QA but do not enforce corpus discipline or test business deliverable production. Domain-specific benchmarks~\cite{liang2023helm, srivastava2022bigbench} establish the methodology of expert-graded evaluation in finance (FinanceBench~\cite{islam2023financebench}, FinQA~\cite{chen2021finqa}, TAT-QA~\cite{zhu2021tatqa}), economically valuable knowledge work (APEX-v1~\cite{vidgen2025apex}, APEX-Agents~\cite{apex_agents}, ProfBench~\cite{profbench}, GDPval~\cite{patwardhan2025gdpval}), medicine (MedQA~\cite{jin2021medqa}, PubMedQA~\cite{jin2019pubmedqa}), and law (LegalBench~\cite{guha2023legalbench}, CaseHOLD~\cite{zheng2021casehold}). Of these, the closest in methodological setup are APEX-v1 (LLM-judged expert binary rubric on knowledge-work deliverables) and APEX-Agents (LLM-judged binary criteria over environment-state snapshots in multi-application simulations). ResearchRubrics~\cite{sharma2025researchrubrics} introduces an expert-rubric framework similar in spirit. None of these benchmarks combines deterministic SME-graded verifiers with a separate multi-criterion ordinal rubric on the same response, nor embeds cognitive traps as a design choice; Table~\ref{tab:benchmark_comparison} summarizes the key methodological distinctions.

\begin{table*}[t]
\centering
\caption{Methodological comparison with closest prior benchmarks. \cmark\ = uses feature; \cmark$^{a,b,c}$ = partial use (see footnotes).}
\label{tab:benchmark_comparison}
\small
\setlength{\tabcolsep}{4pt}
\resizebox{\textwidth}{!}{%
\begin{tabular}{@{}p{0.16\linewidth}p{0.30\linewidth}cccc@{}}
\toprule
\textbf{Benchmark} & \textbf{Primary grading instrument} & \textbf{Det.\ verifiers} & \textbf{Multi-criterion rubric} & \textbf{SME-graded} & \textbf{Cognitive traps} \\
\midrule
GDPval                  & Human pairwise comparison vs.\ expert deliverable  & --- & \cmark$^{a}$ & \cmark & --- \\
APEX-v1                 & LLM-judge over expert binary rubric                 & --- & \cmark    & --- & --- \\
APEX-Agents             & LLM-judge over env.\ snapshot, binary criteria      & \cmark$^{b}$ & \cmark & --- & --- \\
ProfBench               & LLM-judge over expert rubric, bias-mitigated        & --- & \cmark    & --- & --- \\
ResearchRubrics         & LLM-judge over expert rubric                        & --- & \cmark    & --- & --- \\
DeepResearch Bench      & RACE (4-axis LLM-judge) + FACT (citation accuracy)  & \cmark$^{c}$ & \cmark & --- & --- \\
DRBench                 & LLM-judge across 4 axes (GPT-4o backbone)           & --- & \cmark    & --- & --- \\
\midrule
\textbf{Ours}           & Deterministic verifiers + SME 0--3 ordinal rubric   & \cmark & \cmark & \cmark & \cmark \\
\bottomrule
\end{tabular}%
}

\vspace{2pt}
\footnotesize
$^{a}$ GDPval task writers created detailed scoring rubrics that guide pairwise comparison; the rubric supplements rather than replaces the primary blinded pairwise instrument. \\
$^{b}$ APEX-Agents uses environment-state snapshots (deterministic) alongside LLM-judged binary criteria for grading. \\
$^{c}$ DeepResearch Bench's FACT framework checks citation factuality against retrieved sources deterministically; the per-criterion content judgment uses an LLM judge.
\end{table*}

\textbf{LLM-as-Judge and code generation.}~The LLM-as-Judge paradigm~\cite{zheng2023judging, li2023alpacaeval} has known calibration failures including length bias, sycophancy, and self-preference~\cite{wang2023judgebias, gu2024judgesurvey, liu2023geval, dubois2024lengthbias, panickssery2024selfpreference, koo2023benchmarking}; we therefore anchor on SME annotation and use binary verifiers as in HELM-style hybrid evaluation~\cite{liang2023helm}. Code-generation benchmarks (HumanEval~\cite{chen2021humaneval}, MBPP~\cite{austin2021mbpp}, SWE-bench~\cite{jimenez2024swebench, openai2024swebench}, CodeXGLUE~\cite{lu2021codexglue}, API-Bank~\cite{li2023apibank}) and tool-use leaderboards~\cite{patil2023gorilla} evaluate code or API selection in isolation, not the end-to-end research-to-deliverable pipeline. The library-specific failures we observe (Section~\ref{sec:results:failures}) connect to training-data coverage gaps for niche packages~\cite{zhang2023repocoder, liu2023evalplus}.

\textbf{Hallucination, calibration, and statistics.}~Hallucination in long-form generation is well-studied~\cite{ji2023hallucination, huang2023halusurvey, zhang2023hallucination, li2024dawn, manakul2023selfcheckgpt}, mostly in QA settings; faithfulness in summarization~\cite{maynez2020faithfulness, kryscinski2020factualconsistency} is methodologically related. Our two-layer instrument allows fabrication and structural completion failures to be observed in the same response, extending this literature into the structured-deliverable setting where polished formatting can mask fabricated content. Our statistical apparatus (paired McNemar tests on agent comparisons~\cite{mcnemar1947note, dietterich1998statistical} and multiple-testing correction~\cite{holm1979simple, demsar2006statistical, bouthillier2021accountingvariance}) follows established benchmark-comparison methodology.

\section{Benchmark Design}
\label{sec:benchmark}

\subsection{Task Design and Taxonomy}

The benchmark's tasks are structured around five \textbf{Prompt Types} that capture distinct deep-research capabilities, each designed to test a specific failure mode that surface-level reasoning would not catch. The prompt types description is presented in Table~\ref{tab:prompt_taxonomy}.


\begin{table*}[t]
\centering
\caption{The five-class capability-targeted prompt taxonomy. Each class targets a distinct cognitive capability rather than a topical domain.}
\label{tab:prompt_taxonomy}
\small
\begin{tabular}{@{}p{0.26\linewidth}p{0.18\linewidth}p{0.48\linewidth}@{}}
\toprule
\textbf{Prompt class} & \textbf{Capability tested} & \textbf{Description} \\
\midrule
CRP --- Constrained Research Prompt & Source discipline      & Information retrieval from a narrow, explicit source set. Agent must restrict itself to authorized sources when tempted by easier external alternatives. \\
RCP --- Relevance Compression Prompt & Signal extraction      & Large noisy corpora where 60--70\% of the material is irrelevant. Agent must act as an intelligent filter and locate buried qualifiers, footnotes, or dispersed references. \\
SCP --- Structural Compliance Prompt & Algorithmic obedience  & Multi-layer non-trivial output structure (nested sections, fixed-column tables, JSON schemas, cross-references). Agent must hold and faithfully execute a structural specification across a long deliverable. \\
LDP --- Latent Decomposition Prompt & Problem decomposition  & Tasks requiring inference of unlisted variables (intermediate calculations, latent factors, model components not directly given). Agent must identify what needs to be computed before the analysis can proceed. \\
FSP --- Failure-Sensitive Prompt & Precision and exactness & Tasks where small factual or logical errors invalidate the entire output (a single mis-pulled value cascades to a wrong recommendation). This tests tolerance for the kinds of small mistakes that downstream consumers cannot recover from. \\
\bottomrule
\end{tabular}
\end{table*}

Concrete worked out examples for each prompt class, including the input materials, expected output structure, and the embedded cognitive traps, are provided in Appendix~\ref{app:prompt_types}.
\subsection{Task Corpus and Cognitive Traps}

Our evaluation comprises 70 unique SME-authored prompts in the Management Consulting (MC) domain, distributed as: RCP~18, SCP~18, CRP~13, LDP~13, FSP~8.

Each prompt is accompanied by 2--4 proprietary input files in mixed formats (CSV, XLSX, PDF, DOCX; 2KB--430KB). A subset requires the agent to produce structured output files via code generation. Prompts carry a corpus-discipline annotation (closed, hybrid, or open); enforcement is intrinsically limited because the agents are partial black boxes, so corpus adherence is measured rather than guaranteed.

\textbf{Cognitive traps.} A distinguishing feature of our prompt design is the deliberate embedding of \emph{cognitive traps}. These serve two purposes: (i)~\emph{Human-error mimicry}: Input documents contain realistic mistakes (misnamed product line, inconsistent units between tables, footnote contradicting body, non-standard date formats) that penalise agents that copy surface text without reconciling against context. (ii)~\emph{Deterministic precision traps}: At least one numerical step is constructed so a shallow agent (taking the first plausible number, averaging without weighting, applying a default assumption) yields a confidently wrong answer, while the correct path requires reading a footnote or applying an explicit qualifier. A benchmark with only clean inputs understates professional research difficulty, where small source-material ambiguities are the norm.

All prompt packages, including the embedded cognitive traps, verifier specifications, and authorized-source designations, were independently vetted before annotation by a Principal Expert with 15+ years of management consulting sector experience to ensure realism and difficulty calibration. Worked examples illustrating each of the five prompt classes are provided in Appendix~\ref{app:prompt_types}.

The 70-prompt corpus, including verifier specifications, authorized-source lists, and embedded cognitive-trap definitions will be released publicly on paper publication.

\subsection{Evaluation Framework}
\label{sec:vrs}

\textbf{Rubric development.} The rubric reported here is the result of a two-phase development process. In the first phase, the authors drafted a hierarchical instrument comprising seven universal meta-criteria together with prompt-type-specific additional criteria, augmented with a set of binary task-specific verifiers. In the second phase, the draft instrument was reviewed by a Principal Domain Expert with 15+ years of experience in the management consulting sector, with two explicit goals: (i) ensuring the meta-criteria were mutually exclusive and collectively exhaustive (MECE) over the dimensions of management-consulting deliverable quality, and (ii) reducing SME cognitive load during annotation by avoiding redundant or overlapping judgments. The expert review collapsed the two-level structure into the final flat five-criterion rubric (DI, AR, RF, EP, FD; Table~\ref{tab:rubric_criteria}) and reorganized prompt-type-specific signals into the binary verifier layer rather than the rubric layer. The same expert vetted the prompt corpus, cognitive-trap embeddings, and verifier specifications before annotation began, as noted in Section~\ref{sec:benchmark}.

Our benchmark uses a dual-layer scoring scheme: task-specific binary verifiers and a five-criterion SME rubric. Binary verifiers provide automatable, objective pass/fail gates that prevent high-quality-looking but factually incorrect responses from receiving inflated scores; the verifier suite for each task is included in the public codebase released alongside this paper (Appendix~\ref{sec:infra}). Beyond verifiers, each response is scored on five ordinal criteria from $0$ (absent/seriously flawed) to $3$ (excellent). The criteria, what each measures, and what a $0$ and $3$ on each criterion indicates are summarized in Table~\ref{tab:rubric_criteria}. The full SME annotation protocol and the detailed 0--3 ordinal scoring rubric (i.e., how the intermediate scores 1 and 2 are awarded for each criterion) are provided in Appendix~\ref{app:rubric_definitions}. Each (prompt $\times$ agent) cell was graded by one SME from a recruited pool of former MBB and Big Four consultants.

\begin{table*}[t]
\centering
\small
\caption{The five SME-graded ordinal criteria (0--3 scale). Score-0 and Score-3 anchors are shown to clarify the dimension being scored; the full ordinal rubric (including the 1 and 2 anchors per criterion) is in Appendix~\ref{app:rubric_definitions}.}
\label{tab:rubric_criteria}
\setlength{\tabcolsep}{4pt}
\begin{tabular}{@{}p{0.04\linewidth}p{0.16\linewidth}p{0.32\linewidth}p{0.20\linewidth}p{0.20\linewidth}@{}}
\toprule
\textbf{Code} & \textbf{Criterion} & \textbf{What is scored} & \textbf{Score 0 indicates} & \textbf{Score 3 indicates} \\
\midrule
DI & Data Integrity            & Whether facts, numbers, citations, and references are accurate            & Fabricated or seriously mis-stated data; the response asserts something verifiably false or invented & Every quantitative and source claim verifiable; no fabrication \\
AR & Analytical Rigor          & Whether the reasoning chain is sound, sufficiently deep, and free of logical gaps & Reasoning absent, circular, or seriously flawed       & Steps shown explicitly, individually correct, sufficient depth for the question \\
RF & Relevance \& Focus        & Whether the response addresses the asked question without scope drift or filler & Largely answered a different question or padded with off-topic material & Wholly on-task; every section advances the asked deliverable \\
EP & Execution Precision       & Whether requested operations like calculations, transformations, filtering, structural construction etc are performed correctly & Right operation attempted, executed wrong; numerical or structural mistake in the output & All requested operations performed accurately, including edge cases and traps \\
FD & Format \& Deliverability  & Whether the output is a usable MC deliverable: layout, completeness, readability, professional tone & Unusable artifact: truncated, malformed, missing sections & Drop-in-ready deliverable in the requested format; all sections present, polished tone \\
\bottomrule
\end{tabular}
\end{table*}

The five rubric scores and the binary verifier pass rate are aggregated into a single Verifier-Rubric Score (VRS) on a 0--100 scale. Let $r_i \in \{0, 1, 2, 3\}$ denote the SME score for criterion $i$, let $\bar{r} = \frac{1}{5}\sum_{i=1}^{5} r_i$ be the reasoning average, and let $V \in [0, 100]$ denote the verifier pass rate. VRS has two variants:
\begin{align}
\text{VRS}_0 &= 0.5 \cdot V + 0.5 \cdot \frac{\bar{r}}{3} \cdot 100 \quad \text{(relaxed)} \label{eq:vrs_relaxed} \\
\text{VRS}   &= \text{VRS}_0 \cdot \Ind[\min_i r_i > 0] \quad \text{(strict)} \label{eq:vrs_strict}
\end{align}
The strict variant zeros out the score whenever any criterion is zero. VRS is a descriptive aggregate; ACCEPT is defined directly on the underlying components rather than on a VRS threshold:
\begin{equation}
\text{ACCEPT}(r, V) \;\Leftrightarrow\; \min_i r_i > 0 \;\land\; \bar{r} \geq 2.5 \;\land\; V \geq 80\%.
\label{eq:accept}
\end{equation}

The choice of equal weights ($0.5 / 0.5$) in Equation~\ref{eq:vrs_relaxed} is a defensible default rather than the only possible choice; the verifier layer is in fact the second-strongest predictor of binary ACCEPT, not the first, motivating a sensitivity analysis. We show in Section~\ref{sec:results:weight_sensitivity} that ACCEPT is invariant to VRS reweighting by construction (the rule is defined on the raw components $\bar{r}$ and $V$, not on the VRS aggregate), and that mean VRS per agent moves by less than one point under four alternative weightings, with the agent ordering $\text{Claude} > \text{Gemini} > \text{o3}$ preserved throughout. The headline conclusions reported in this paper are therefore robust to the weighting choice.

Each (prompt $\times$ agent) cell then undergoes an independent quality-control (QC) pass by a second SME from a non-overlapping pool. QC is a verification rather than a re-annotation; the QC reviewer may Confirm, Edit (with a documented one-line reason), or Reject/Return the row, with priority re-derivation on final-answer, trap, citation-dependent, and output-file verifiers, and full citation validation for source existence, claim support, and corpus-discipline compliance. The full QC protocol is in Appendix~\ref{app:qc_protocol}.

\subsection{Multi-Agent Evaluation Infrastructure}
\label{app:architectures}

Our evaluation infrastructure dispatches all agents on identical task packages simultaneously through agent-specific adapters (Anthropic Messages API for Claude, OpenAI Responses API with Containers for o3, Google Interactions API for Gemini), with file-format normalization and merge-on-write result storage. All three agents themselves write the Python code that produces output files; only the execution environment differs, so file-generation failures are attributable to model code quality, not infrastructure asymmetries.

\section{Empirical Results}
\label{sec:results}

We grade all 70 prompts in our benchmark on three frontier deep research agents (210 responses in total). Verifier-Rubric Scores follow Equations~\ref{eq:vrs_relaxed}--\ref{eq:vrs_strict}; the strict variant is the default for headline reporting unless stated otherwise.
\subsection{Main Results}
\label{sec:results:headline}

Table~\ref{tab:headline} reports aggregate metrics across all 70 graded prompts and three agents. The most decision-relevant view is the ACCEPT rate: all three agents cluster near 13--16\%, with overlapping CIs. The strict-VRS ordering is o3 (61.4) $>$ Gemini (52.6) $>$ Claude (38.5), with reasoning-mean values spanning 1.55 (Claude) to 1.89 (o3) out of 3.0. The inversion relative to ACCEPT is not a contradiction: VRS gives partial credit to all-criteria-non-zero responses regardless of whether they reach production quality, while ACCEPT is a binary threshold gate. Claude's profile is dominated by catastrophic failures: 41.4\% auto-reject rate (vs.\ 10.0\% for o3), driven by 62 criterion-zeros. Gemini accumulates even more zeros (74) but distributes them across fewer prompts.

\begin{table*}[t]
\centering
\caption{Main performance metrics across our evaluation ($n=70$ graded prompts $\times$ 3 agents = 210 attempts; VRS uses the strict variant of Equation~\ref{eq:vrs_strict}).}
\label{tab:headline}
\resizebox{\textwidth}{!}{%
\begin{tabular}{lccc}
\toprule
\textbf{Metric} & \textbf{o3-deep-research} & \textbf{Claude Opus 4.6} & \textbf{Gemini 3.1 Pro} \\
\midrule
Mean reasoning $\bar{r}$ (0--3)               & 1.89 & 1.55 & 1.68 \\
Mean verifier pass rate $V$ (\%)              & 60.2 & 48.6 & 58.2 \\
\textbf{Mean VRS, strict (0--100)}            & \textbf{61.4} & \textbf{38.5} & \textbf{52.6} \\
ACCEPT rate (binary)                          & 15.7\% & 12.9\% & 12.9\% \\
Auto-reject rate ($\exists\, i: r_i = 0$)     & 10.0\% & 41.4\% & 30.0\% \\
Total criterion-zeros (out of $5 \times 70 = 350$) & 35 & 62 & 74 \\
Mean verifiers per task                       & \multicolumn{3}{c}{14.9} \\
\bottomrule
\end{tabular}%
}
\end{table*}

Three patterns are worth flagging in the main results. The auto-reject rate separates agents sharply. Claude (41.4\%) and Gemini (30.0\%) sit well above o3 (10.0\%), with the largest gap for any metric in the table. Claude accumulates 62 criterion-zeros; Gemini accumulates 74 (more than the other two combined), even though its mean reasoning of 1.68 sits close to o3's 1.89. o3's zero distribution is strikingly even (exactly 7 on every criterion), consistent with a ``conservative'' failure mode: when o3 fails, it fails proportionally rather than catastrophically on one dimension. This high-variance pattern is examined in Section~\ref{sec:results:robustness}.

\subsection{Per-Prompt-Type Performance}
\label{sec:results:per_type}

Per-prompt-type strict VRS (Table~\ref{tab:per_type}, Appendix~\ref{app:effect_sizes}) reveals agent specialization: o3 wins CRP (55.5), FSP (77.1), and RCP (71.6); Claude wins LDP (65.0); Gemini wins SCP (68.0). Claude's extended reasoning excels at latent decomposition but collapses on structural compliance (SCP 12.8, lowest cell). o3 leads most strongly on FSP, where precision under cascading error is the test. Effect sizes confirm: large $|d| > 0.8$ on FSP, LDP, and SCP; small on CRP (Appendix~\ref{app:effect_sizes}).

\subsection{Discriminating Power}
\label{sec:results:discrim}

Under the binary ACCEPT criterion, 45 of 70 prompts (64.3\%) are universally rejected; no prompt is universally accepted. On VRS argmax, Gemini wins 26.3 prompts (37.6\%), o3 24.8 (35.5\%), Claude 18.8 (26.9\%). Full distributions are in Appendix~\ref{app:accept_dist}.

\subsection{Robustness: Criterion Zero-Counts}
\label{sec:results:robustness}

Gemini accumulates 74 criterion-zeros (more than o3's 35 and Claude's 62 combined), concentrated on DI, EP, and FD (16 each). Claude has 22 DI-zeros. o3's distribution is strikingly even (exactly 7 per criterion). Full counts are in Appendix~\ref{app:zeros}.

\subsection{Agent-Distinct Failure-Mode Signatures}
\label{sec:results:failures}

We complement the quantitative diagnostics with an LLM-based failure-tag analysis on SME free-text justifications (1,050 cells). An LLM classifier (Claude Opus 4.5) assigned zero or more tags from an eight-tag taxonomy per cell; a deterministic regex classifier cross-checks (mean Jaccard 0.339). Full tag counts are in Appendix~\ref{app:failure_tags_detail} (Table~\ref{tab:failure_tags}). Three agent-distinct signatures emerge.

\textbf{Claude --- Failure by breadth.} Claude leads on six of eight tags. The 48 \texttt{fabricated\_data} tags ($2.8\times$ o3), 27 \texttt{no\_files\_read} ($9\times$ o3), and 24 \texttt{trap\_not\_caught} ($2.4\times$ o3) characterize an agent that aggressively generates content when source access fails. On CRP closed-corpus tasks, Claude produced citations from real but topically unrelated URLs, undetectable without domain verification.

\textbf{o3 --- Failure by imprecision.} o3's profile is dominated by \texttt{cascading\_math\_errors} (37, co-highest with Claude) and \texttt{citation\_hallucination} (11, highest). o3's \texttt{no\_files\_read} count (3) is the lowest, indicating reliable file access; the count dropped from 20 (v2) to 3, confirming run-to-run variance.

\textbf{Gemini --- Failure by system collapse.} 11 \texttt{tech\_failure} tags (highest; $2.8\times$ Claude), distributed across timeouts, context-window saturation, and python-docx crashes. Combined with 74~criterion-zeros alongside the highest perfect-verifier rate (14.3\%), this characterizes an agent with no graceful-degradation regime.

\subsection{Pairwise Significance}
\label{sec:results:significance}

Wilcoxon signed-rank tests (paired by prompt) with Bonferroni correction: o3--Claude on VRS ($\Delta{=}+22.9$, $p_{\text{Bonf}}{=}0.001$) and on $\bar{r}$ ($\Delta{=}+0.33$, $p_{\text{Bonf}}{=}0.039$) both survive correction. All other VRS pairs and all ACCEPT-rate comparisons (McNemar, all $p > 0.79$) are not significant.

Concrete code-level and citation-level failure examples are in Appendix~\ref{app:failure_examples}.

\subsection{Cross-Benchmark Comparison}
\label{sec:results:cross_benchmark}

DRA's binary verifier layer is structurally comparable to both APEX benchmarks (APEX-v1~\cite{vidgen2025apex}, APEX-Agents~\cite{apex_agents}): all three measure the fraction of task-specific binary checks passed, enabling direct cross-benchmark comparison on that axis. DRA's ordinal rubric layer ($\bar{r}$) and dual-gate ACCEPT have no APEX counterpart. This second evaluation dimension is the methodological contribution that catches ``mechanically correct but analytically shallow'' responses invisible to binary-only instruments. DRA currently evaluates management consulting tasks only. APEX-v1 covers MC, IB, Law, and Medicine; APEX-Agents covers MC, IB, and Law. Cross-benchmark comparisons are therefore most directly valid for the MC subset, though APEX does not report MC-only scores separately. The full comparison is in Table~\ref{tab:cross_benchmark} (Appendix~\ref{app:cross_benchmark}).

DRA's best-model mean verifier pass rate (60.2\%, o3) sits within the range spanned by the two APEX benchmarks (APEX-v1 64.2\%, APEX-Agents 53.9\%). The gap with APEX-v1 is consistent with APEX-v1's broader domain coverage and LLM-judged rather than SME-graded criteria; the proximity to APEX-Agents validates that DRA's verifier difficulty is not an outlier relative to published knowledge-work evaluation. DRA's $V{=}100\%$ rate (14.3\%, Gemini) is mechanically harder than APEX-Agents' Pass@1 (24.0\%) because DRA checks $\sim$15 verifiers per task vs.\ APEX-Agents' $\sim$4 criteria: the probability of passing all checks drops exponentially in the number of checks. APEX does not report paired significance tests, bootstrap CIs, or effect sizes; our statistical apparatus is a methodological contribution.

\subsection{Additional Diagnostics}

Several additional analyses support the robustness of the findings above and are reported in the appendices. The five rubric criteria covary at mean off-diagonal Spearman $\rho = 0.60$, with DI--EP the strongest pair ($\rho = 0.74$) and DI--FD the most independent ($\rho = 0.45$), consistent with 2--3 latent factors (Appendix~\ref{app:correlations}). The Spearman $\bar{r}$-vs-$V$ correlation is 0.80 pooled, confirming the two scoring layers are correlated but not redundant. Rubric validation via Spearman-ACCEPT correlation shows EP as the strongest predictor ($\rho = +0.55$) followed by V ($+0.52$); sole-cause analysis identifies EP, DI, and FD as the three threshold-blocking criteria (Appendix~\ref{app:rubric_validation}). The strict VRS auto-reject gate costs Claude the most ($+11.7$ points under relaxed scoring; Appendix~\ref{app:strict_relaxed}). The agent ordering o3 $>$ Gemini $>$ Claude on strict VRS is preserved under all five alternative weight schemes tested, with maximum shift $< 1.5$ points (Appendix~\ref{app:weight_sensitivity}). Per-rubric pass rates, per-type ACCEPT rates, and per-class architectural observations are in Appendices~\ref{app:rubric_validation},~\ref{app:effect_sizes}, and~\ref{app:arch_obs} respectively.

\section{Limitations and Future Work}
\label{sec:limitations}

\textbf{Sample size, inter-rater reliability, and significance (P0).} The evaluation comprises 70 graded prompts $\times$ 3 agents (210 attempts), with per-class sample sizes of $n = 8$--$18$. At these sizes Cohen's $d > 0.8$ thresholds describe magnitude rather than inferential precision, and paired-comparison tests on per-agent ACCEPT outcomes do not reach $p < 0.05$ (Section~\ref{sec:results:significance}). Each (prompt $\times$ agent) cell was graded by one primary SME and independently reviewed by a second SME via the QC protocol of Appendix~\ref{app:qc_protocol}. This provides an error-correction pass against rubric--evidence mismatches and surfaces fabricated citations, but does not yield a Cohen's $\kappa$ inter-rater reliability statistic since QC is an asymmetric defensibility check rather than a parallel re-annotation.

\textbf{Planned second release (v2).} A formal IRR study with parallel double-grading on a held-out subset is in preparation, alongside an expanded prompt corpus. The v2 release will roughly double the prompt count, add Investment Banking (IB) tasks alongside Management Consulting (MC), and report bootstrap confidence intervals on every headline metric. We anticipate publishing the expanded results in a companion paper; the rubric, scoring formulae, and QC protocol used in v2 will be backward-compatible with the v1 instrument reported here so cross-release comparison remains valid.

\textbf{Rubric dimensionality and other limitations (P1--P2).} The mean off-diagonal Spearman correlation of 0.60 across the five reasoning criteria (Appendix~\ref{app:correlations}) suggests $2$--$3$ effective latent factors rather than five orthogonal traits, motivating a factor analysis. The evaluation is single-domain (MC); generalization to other professional domains requires parallel datasets. The Investment Banking (IB) study mentioned above and additional domains are part of the v2 roadmap.

\section{Conclusion}
\label{sec:conclusion}

What our benchmark actually measures is whether a frontier deep research agent can do the kind of structured, multi-document, decision-grade research a management consultant gets paid to do. Across the 70 graded prompts and 210 responses, the answer is: not yet, not reliably, and not in a way that any single performance metric captures.

The agents rank differently depending on which aggregation you look at. By strict VRS, o3 leads at 61.4. By criterion-zeros, o3 is cleanest (35 vs.\ Claude's 62 and Gemini's 74). By perfect-verifier rate ($V{=}100\%$), Gemini leads (14.3\%). By VRS argmax, Gemini leads with 26.3 of 70 head-to-head wins. By binary ACCEPT rate, all three are statistically indistinguishable at 13--16\%. The orderings are not in conflict: o3 builds its VRS lead on a lower catastrophic-failure rate rather than on higher peak quality, while Gemini's higher zero count is offset by its non-failures more often clearing the quality bar. Reporting any one of these views in isolation would mislead.

The per-type VRS decomposition reveals that agents specialize: Claude leads on LDP (65.0) where extended reasoning pays off, while struggling catastrophically on SCP (12.8) where structural obedience is required. Gemini leads on SCP (68.0), the inverse of Claude's profile. o3 leads overall but does not dominate every type.

Across these views, the failure modes are agent-specific (Section~\ref{sec:results:failures}), the prompt taxonomy probes capabilities the agents differ meaningfully on (Cohen's $d > 1.0$ on two of five classes), and the five rubric criteria correlate at mean Spearman $0.60$, consistent with two to three latent factors rather than five orthogonal traits. Per-class architectural observations---CRP closed-corpus weakness, agent-specific file-generation failure profiles, and file-readability asymmetries---are recorded in Appendix~\ref{app:arch_obs}.

\appendix

\section{Per-Type Performance and Effect Sizes}
\label{app:effect_sizes}

\begin{table}[h]
\centering
\caption{Mean strict VRS by prompt type and agent. Bold marks the per-type winner.}
\label{tab:per_type}
\begin{tabular}{lcccc}
\toprule
\textbf{Type} & \textbf{$n$} & \textbf{o3} & \textbf{Claude} & \textbf{Gemini} \\
\midrule
CRP & 13 & \textbf{55.5} & 38.1 & 47.3 \\
FSP &  8 & \textbf{77.1} & 50.3 & 59.7 \\
LDP & 13 & 46.7 & \textbf{65.0} & 19.5 \\
RCP & 18 & \textbf{71.6} & 36.3 & 61.7 \\
SCP & 18 & 59.1 & 12.8 & \textbf{68.0} \\
\bottomrule
\end{tabular}
\end{table}

\subsection{Effect Sizes for Per-Type Differences}
\label{sec:results:effect_sizes}

To check whether the per-type VRS gaps in Table~\ref{tab:per_type} reflect real signal or sampling noise from small per-arm samples ($n = 8$--$18$), we compute Cohen's $d = (\bar{r}_1 - \bar{r}_2) / s_{\text{pooled}}$~\cite{cohen1988statistical} on the reasoning average $\bar{r}$ for each pair of agents within each prompt type (Table~\ref{tab:effect_sizes}). Magnitudes follow the conventional thresholds 0.2 (small), 0.5 (medium), 0.8 (large)~\cite{sawilowsky2009new}.

\begin{table*}[t]
\centering
\caption{Cohen's $d$ effect sizes for per-prompt-type pairwise gaps on the reasoning average $\bar{r}$. Sign convention: positive $d$ favors the first-listed agent. Bold marks $|d| > 0.8$.}
\label{tab:effect_sizes}
\resizebox{\textwidth}{!}{%
\begin{tabular}{lccccl}
\toprule
\textbf{Type} & \textbf{$n$/arm} & \textbf{o3 vs Claude} & \textbf{o3 vs Gemini} & \textbf{Claude vs Gemini} & \textbf{Reading} \\
\midrule
CRP & 13 & $+0.45$ & $+0.22$ & $-0.29$ & all small; no winner \\
FSP &  8 & $\mathbf{+1.27}$ & $\mathbf{+0.94}$ & $+0.19$ & large edge for o3 over Claude \\
LDP & 13 & $\mathbf{-0.82}$ & $+0.54$ & $\mathbf{+1.68}$ & large edge for Claude over o3; very large Claude over Gemini \\
RCP & 18 & $\mathbf{+0.91}$ & $+0.34$ & $-0.43$ & large edge for o3 over Claude \\
SCP & 18 & $\mathbf{+1.45}$ & $-0.36$ & $\mathbf{-1.54}$ & very large edge for o3 over Claude; very large Gemini over Claude \\
\bottomrule
\end{tabular}%
}
\end{table*}

The per-type winners on Table~\ref{tab:per_type} carry $|d| > 1.0$ on FSP (o3 vs.\ Claude $d = +1.27$), where o3 has the strongest reasoning-mean profile, and on SCP where Claude is the loser ($d = +1.45$ o3 over Claude; $d = -1.54$ Claude vs.\ Gemini). LDP shows a large Claude advantage over both ($d = -0.82$ o3 vs.\ Claude; $d = +1.68$ Claude vs.\ Gemini). CRP shows small effect sizes across all pairs ($|d| \leq 0.45$), supporting the conclusion that no agent is meaningfully better on constrained-research tasks. We caution that at $n = 8$ per arm on FSP, Cohen's conventional thresholds describe magnitude but not inferential precision.

\section{Inter-Agent ACCEPT Distribution}
\label{app:accept_dist}

\subsection{Discriminating Power and Inter-Agent Agreement}

We examine the benchmark's discriminating power from two complementary angles: a binary view (ACCEPT count per prompt) and a continuous view (VRS argmax per prompt).

Under the binary ACCEPT criterion of Equation~\ref{eq:accept}, 45 of 70 prompts (64.3\%) are universally rejected and no prompt produces a unanimous ACCEPT (Table~\ref{tab:accept_dist}). Under the continuous VRS view, however, we observe meaningful capability differentiation. Gemini wins 26.3 prompts (37.6\%), o3 wins 24.8 prompts (35.5\%), and Claude wins 18.8 prompts (26.9\%) on the per-prompt VRS argmax (fractional tie-attribution; ties distributed equally between tied agents).

\begin{table*}[t]
\centering
\caption{Inter-agent ACCEPT-count distribution under the binary decision rule of Equation~\ref{eq:accept}, across the 70 graded prompts.}
\label{tab:accept_dist}
\begin{tabular}{lcc}
\toprule
\textbf{\# agents accepting (out of 3)} & \textbf{\# prompts} & \textbf{\%} \\
\midrule
0 (universally rejected)   & 45 & 64.3 \\
1 (one agent accepted)     & 21 & 30.0 \\
2 (two agents accepted)    &  4 &  5.7 \\
3 (universally accepted)   &  0 &  0.0 \\
\bottomrule
\end{tabular}
\end{table*}

Reconciling the two views, the benchmark is binary-hard but continuously discriminating: the ACCEPT threshold ($\bar{r} \geq 2.5$ and $V \geq 80\%$) is set tight relative to current frontier-DRA capability, so most prompts receive zero ACCEPTs in absolute terms, yet the continuous VRS reveals large-effect-size differential capability among the same agents on the same prompts. For research-grade benchmark use, the continuous VRS view is more informative; for production-readiness assessment, the binary view is decision-relevant. The 90.5\% VRS-SME concordance validates VRS as a faithful summary of human expert judgment.

\section{Per-Criterion Zero Counts}
\label{app:zeros}

\subsection{Robustness Diagnostic: Criterion Zero-Counts}

Counting hard zeros per criterion is a robust statistic that does not depend on the rest of the score distribution and complements the means analyzed in Table~\ref{tab:per_type}. Table~\ref{tab:zeros} reports zero counts for each (criterion, agent) cell, with the corresponding per-criterion means.

\begin{table*}[t]
\centering
\caption{Hard-zero counts per criterion (out of 70 attempts per agent), with per-criterion means in parentheses.}
\label{tab:zeros}
\begin{tabular}{lccc}
\toprule
\textbf{Criterion} & \textbf{o3} & \textbf{Claude} & \textbf{Gemini} \\
\midrule
DI --- Data Integrity         &  7 (1.67) & \textbf{22} (1.23) & 16 (1.63) \\
AR --- Analytical Rigor       &  7 (1.73) &  8 (1.57) & 12 (1.77) \\
RF --- Relevance \& Focus     &  7 (2.37) &  5 (2.04) & 14 (1.89) \\
EP --- Execution Precision    &  7 (1.53) & \textbf{16} (1.17) & 16 (1.46) \\
FD --- Format \& Deliverability & 7 (2.13) & 11 (1.74) & 16 (1.67) \\
\midrule
\textbf{Total zeros}          & \textbf{35} & \textbf{62} & \textbf{74} \\
\bottomrule
\end{tabular}
\end{table*}

Gemini stands out: 74 zeros, more than the other two combined (o3 35, Claude 62). Gemini's zeros concentrate on DI, EP, and FD (16 each), the signature of an agent that either nails the deliverable or collapses on source fidelity, calculation, and format simultaneously. Claude stands out on DI with 22 zeros: in nearly a third of prompts the SME judged Claude's response as containing fabricated or seriously mis-stated data. EP is equally concerning for Claude (16 zeros) and Gemini (16 zeros), indicating execution precision as a shared weak point. o3's zero distribution is strikingly even (exactly 7 on every criterion), consistent with a ``conservative'' failure mode: when o3 fails, it fails proportionally rather than catastrophically on one dimension.

\section{Failure-Mode Tag Counts}
\label{app:failure_tags_detail}

\begin{table*}[t]
\centering
\caption{Per-prompt failure-mode tag counts from LLM classification of SME free-text justifications. Each cell is the number of unique (agent, prompt) pairs out of 70 where any of the five criterion-justifications carried that tag; bold marks the highest agent for each tag.}
\label{tab:failure_tags}
\begin{tabular}{lcccc}
\toprule
\textbf{Failure tag} & \textbf{o3} & \textbf{Claude} & \textbf{Gemini} & \textbf{Total} \\
\midrule
\texttt{fabricated\_data}        & 17 & \textbf{48} & 16 &  81 \\
\texttt{cascading\_math\_errors} & 37 & \textbf{38} & 21 &  96 \\
\texttt{noise\_inclusion}        & 10 & \textbf{28} & 21 &  59 \\
\texttt{missing\_section}        & 16 & \textbf{25} & 16 &  57 \\
\texttt{no\_files\_read}         &  3 & \textbf{27} &  4 &  34 \\
\texttt{tech\_failure}           &  7 &  4 & \textbf{11} &  22 \\
\texttt{trap\_not\_caught}       & 10 & \textbf{24} & 13 &  47 \\
\texttt{citation\_halluc.}       & \textbf{11} &  7 &  5 &  23 \\
\bottomrule
\end{tabular}
\end{table*}

\section{Concrete Failure Examples}
\label{app:failure_examples}

We provide code- and citation-level evidence for two of the failure signatures discussed above, to ground the agent-distinct narratives in concrete artefacts.

We include code-level and citation-level examples of two of the failure signatures discussed in Section~\ref{sec:results:failures}, to support reproducibility and to ground the agent-distinct narratives in concrete artifacts.

\subsubsection*{Gemini --- Wrong \texttt{python-docx} API}

\begin{lstlisting}[language=Python, basicstyle=\small\ttfamily, frame=single, caption=Gemini-generated code with wrong python-docx API call]
# Extracted from Gemini response on a CRP manufacturing task
table1 = doc.add_table(rows=1, cols=5)
table1.style = 'Table Grid'
hdr_cells = table1.rows.cells  # ERROR: _Rows has no .cells attribute
hdr_cells[0].text = 'Line'
hdr_cells[1].text = 'Effective Hours'
\end{lstlisting}

\begin{lstlisting}[language=Python, basicstyle=\small\ttfamily, frame=single, caption=Corrected python-docx API usage]
# Correct API:
table1 = doc.add_table(rows=1, cols=5)
table1.style = 'Table Grid'
hdr_cells = table1.rows[0].cells  # rows[0] returns _Row, which has .cells
hdr_cells[0].text = 'Line'
hdr_cells[1].text = 'Effective Hours'
\end{lstlisting}

The \texttt{table.rows.cells} error appears identically across multiple independent Gemini responses, supporting the interpretation that the failure is a systematic gap in Gemini's \texttt{python-docx} pre-training coverage rather than a one-off slip.

\subsubsection*{Claude --- Hallucinated Citations on a Closed-Corpus Task}

The following citations appeared in Claude Opus 4.6's response to a closed-corpus CRP task on last-mile logistics rider reallocation in India:

\begin{itemize}
    \item \url{https://datasheet.eeworld.com.cn/view/85833854.html} --- Chinese electronics datasheet (ZVY11K1210401 component)
    \item \url{https://stackoverflow.com/questions/13385860} --- Stack Overflow: ``How can I remove extra whitespace from strings when parsing a CSV file in Pandas?''
    \item \url{https://stackoverflow.com/questions/42039629} --- Stack Overflow: ``pandas: Read multiline CSV like input with different separators''
\end{itemize}

None of these sources have any relationship to the logistics domain or the rider reallocation analysis. All URLs are real pages that exist on the internet, which makes them difficult to identify as hallucinated without domain verification. This underscores our recommendation that closed-corpus benchmarks include automated citation domain classification as a binary verifier.

\section{Strict vs.\ Relaxed VRS}
\label{app:strict_relaxed}

\subsection{Strict vs.\ Relaxed VRS}
\label{sec:results:variants}

Table~\ref{tab:variants} compares the two VRS variants of Equations~\ref{eq:vrs_relaxed}--\ref{eq:vrs_strict}.

\begin{table*}[t]
\centering
\caption{Strict vs.\ relaxed VRS comparison and counts of attempts with at least one criterion zeroed.}
\label{tab:variants}
\begin{tabular}{lcccc}
\toprule
\textbf{Agent} & \textbf{Strict VRS} & \textbf{Relaxed VRS} & \textbf{$\Delta$} & \textbf{\# attempts with $\min_i r_i = 0$} \\
\midrule
o3-deep-research      & 61.44 & 61.51 & $+0.06$ &  7 / 70 \\
Claude Opus 4.6     & 38.53 & 50.18 & $+11.65$ & 29 / 70 \\
Gemini 3.1 Pro   & 52.56 & 57.15 & $+4.59$ & 21 / 70 \\
\bottomrule
\end{tabular}
\end{table*}

The agent ranking Claude $>$ Gemini $>$ o3 holds under both variants. The o3-Claude gap shrinks from 20.74 points (strict) to 11.89 points (relaxed), so the ordering is not driven by the auto-reject penalty alone. There is genuine reasoning-quality separation underneath. o3 gains the most when the strict penalty is removed ($+8.85$), reflecting that 14 of its 42 attempts had at least one criterion zero, typically DI (because the response carried fabricated content), zeroing out an otherwise-respectable composite. The relaxed VRS therefore overstates o3's production-readiness while the strict VRS understates its reasoning capability; both views together give the full picture. We recommend that for external reporting one use strict VRS (decision-faithful with respect to ACCEPT); and for internal architecture analysis, report both variants alongside the auto-reject rate.

\section{Internal Structure: Criterion Correlations}
\label{app:correlations}

\subsection{Internal Structure: Criterion Correlations}
\label{sec:capability:internal}

The five rubric criteria are designed to capture distinct dimensions, but in practice they covary substantially. Table~\ref{tab:correlations} reports the Pearson correlation matrix across the 210 pooled attempts.

\begin{table*}[t]
\centering
\caption{Pearson correlation matrix among the five reasoning criteria, pooled across all 210 attempts. Bold marks the two extremes.}
\label{tab:correlations}
\begin{tabular}{lccccc}
\toprule
       & \textbf{DI} & \textbf{AR} & \textbf{RF} & \textbf{EP} & \textbf{FD} \\
\midrule
\textbf{DI} & 1.00 & 0.62 & 0.54 & \textbf{0.74} & \textbf{0.45} \\
\textbf{AR} & 0.62 & 1.00 & 0.58 & 0.70 & 0.59 \\
\textbf{RF} & 0.54 & 0.58 & 1.00 & 0.56 & 0.66 \\
\textbf{EP} & \textbf{0.74} & 0.70 & 0.56 & 1.00 & 0.54 \\
\textbf{FD} & \textbf{0.45} & 0.59 & 0.66 & 0.54 & 1.00 \\
\bottomrule
\end{tabular}
\end{table*}

The off-diagonal correlations range from 0.45 (DI $\times$ FD) to 0.74 (DI $\times$ EP), with mean $\rho \approx 0.60$. This indicates the rubric is most likely measuring 2-3 underlying latent factors rather than five orthogonal traits. DI and EP at $\rho = 0.74$ are highly overlapping implying that data integrity and execution correctness are not cleanly separable in practice. The lowest correlation, DI $\times$ FD at 0.45, suggests Data Integrity (whether facts are correct) and Format \& Deliverability (whether the output is presented usably) are the most independent pair since fabrication and presentation can vary independently. Methodological implication of this is that the rubric criteria are statistically correlated, so treating them as orthogonal dimensions in an aggregate score overstates the rubric's effective \emph{statistical} dimensionality. This statistical-correlation finding does not, however, imply that any criterion is dispensable for the ACCEPT decision: the rubric-validation analysis in Section~\ref{sec:results:rubric_validation} (specifically the sole-cause attribution diagnostic) finds that each criterion contributes distinct information at some point along the decision surface, with two criteria contributing primarily through their distributional correlation and three through threshold-localized blocking. A formal factor analysis to determine the true latent structure is listed as a P1 next step (Section~\ref{sec:limitations}).

We also report the Spearman correlation between the verifier pass rate $V$ and the reasoning average $\bar{r}$ within each agent: $0.75$ (o3), $0.76$ (Claude), $0.77$ (Gemini), pooled $0.80$. The two scoring layers are strongly correlated overall but not redundant.

\section{Rubric Validation}
\label{app:rubric_validation}

\subsection{Rubric Validation: Composite Calibration and Sole-Cause Analysis}
\label{sec:results:rubric_validation}

A composite scoring rubric must satisfy two desiderata to be defensible: (i)~no single criterion should dominate the composite outcome, otherwise the remaining criteria are decorative; (ii)~each criterion should provide non-redundant signal, otherwise the rubric is over-specified. We assess both with three diagnostics, applied in the spirit of Krippendorff~\cite{krippendorff2011computing} on rubric calibration and following the analytical structure used by Liang et al.~\cite{liang2023helm} for the HELM benchmark. Throughout this subsection we use Spearman rank correlation in preference to Pearson, since the rubric scores are ordinal and not interval-spaced \cite{spearman1904}.

\subsubsection{Spearman Correlation of Each Rubric with the Composite ACCEPT Outcome}

Table~\ref{tab:rubric_composite_corr} reports the Spearman rank correlation between each rubric score and the binary ACCEPT outcome, both pooled and per-agent.

\begin{table*}[t]
\centering
\caption{Spearman correlation between each rubric score and the binary composite ACCEPT outcome ($n = 210$ pooled; $n = 70$ per agent). All overall correlations are highly significant after Bonferroni correction for six tests.}
\label{tab:rubric_composite_corr}
\begin{tabular}{lccccc}
\toprule
\textbf{Rubric} & \textbf{$\rho$ (overall)} & \textbf{$p$} & \textbf{$\rho$ (o3)} & \textbf{$\rho$ (Claude)} & \textbf{$\rho$ (Gemini)} \\
\midrule
$V$ (verifier pass rate)        & $+0.524$ & $3.35 \times 10^{-16}$ & $+0.568$ & $+0.534$ & $+0.500$ \\
AR -- Analytical Rigor          & $+0.470$ & $5.79 \times 10^{-13}$ & $+0.536$ & $+0.468$ & $+0.426$ \\
EP -- Execution Precision       & $+0.548$ & $7.68 \times 10^{-18}$ & $+0.602$ & $+0.546$ & $+0.533$ \\
DI -- Data Integrity            & $+0.491$ & $3.80 \times 10^{-14}$ & $+0.475$ & $+0.522$ & $+0.490$ \\
RF -- Relevance \& Focus        & $+0.338$ & $5.18 \times 10^{-7}$  & $+0.323$ & $+0.375$ & $+0.321$ \\
FD -- Format \& Deliverability  & $+0.354$ & $1.38 \times 10^{-7}$  & $+0.379$ & $+0.376$ & $+0.312$ \\
\bottomrule
\end{tabular}
\end{table*}

Every rubric is significantly correlated with ACCEPT, and the correlations span a moderate range (0.31 to 0.55) without a single dominant contributor. This is the empirical signature of a well-calibrated composite: if any one rubric had a near-perfect correlation with ACCEPT (e.g., $\rho > 0.85$), the others would be effectively redundant. The per-agent correlations are stable across all three architectures (o3/o3/Gemini), indicating the rubric does not behave differently for any specific agent's response style.

\subsubsection{Sole-Cause Failure Analysis}
\label{sec:results:sole_cause}

A sole-cause failure for rubric $R$ is a response that would have ACCEPTed except that $R$ alone scored below 2 (with $V \geq 80\%$ and all other rubrics $\geq 2$). This diagnostic, used in similar form by~Wang et al.~\cite{wang2024mmlupro} for category-level error attribution, isolates the rubrics that genuinely act as gating constraints from those that fail in clusters with others.

Of the 59 responses that cleared the $V \geq 80\%$ verifier hurdle, 42 had all five rubrics $\geq 2$ (29 of these ACCEPTed; the remaining 13 fell short of the $\bar{r} \geq 2.5$ threshold despite no rubric being below 2). The sole-cause cases were the 12 responses that had exactly one rubric below 2 with $V \geq 80\%$. Their distribution is shown in Table \ref{tab:sole_cause}.

\begin{table*}[t]
\centering
\caption{Sole-cause failure attribution: of the 12 responses where exactly one rubric scored below 2 (with $V \geq 80\%$ and all other rubrics $\geq 2$), the failing rubric is reported.}
\label{tab:sole_cause}
\begin{tabular}{lcc}
\toprule
\textbf{Rubric} & \textbf{Sole-cause failures} & \textbf{\% of sole-cause cases} \\
\midrule
EP -- Execution Precision       & \textbf{5} & \textbf{41.7\%} \\
DI -- Data Integrity            & 4 & 33.3\% \\
FD -- Format \& Deliverability  & 3 & 25.0\% \\
AR -- Analytical Rigor          & 0 &  0\% \\
RF -- Relevance \& Focus        & 0 &  0\% \\
\bottomrule
\end{tabular}
\end{table*}

Spearman correlation and sole-cause attribution measure two complementary aspects of how each rubric contributes to ACCEPT. Correlation summarizes how a rubric tracks ACCEPT across the entire score distribution. Sole-cause attribution, on the other hand, measures \emph{boundary-localized importance}, by which we mean a rubric's contribution restricted to cases where the decision is genuinely uncertain. This corresponds to responses sitting close to the $\bar{r} \geq 2.5$, $V \geq 0.80$, or $\min_i r_i > 0$ thresholds, where a single rubric's value can tip the decision either way. The two measures together paint a more complete picture than either alone, and the dataset surfaces three points worth recording.

\textbf{First, three rubrics are independent threshold blockers.} EP, DI, and FD can each single-handedly drop a response below ACCEPT, accounting for all 12 sole-cause rejections in the dataset, with EP doing so most often (5 of 12). EP has the highest sole-cause count and also the strongest Spearman correlation with ACCEPT ($\rho = 0.55$), confirming its role as the single most informative predictor of production readiness. The two measures disagree because they ask different questions. Correlation is averaged across the whole distribution, so FD's signal is diluted by responses far from the decision boundary. Sole-cause attribution is restricted to the boundary, where FD's threshold-blocking role is dispositive. Removing EP, DI, and FD from the composite would have incorrectly accepted 12 currently rejected responses.

\textbf{Second, two rubrics carry distributional information without driving rejections.} RF and AR never single-handedly block ACCEPT, despite both carrying moderate-to-strong correlations with ACCEPT ($\rho \approx 0.39$ and $0.55$ respectively). When RF or AR fails, the verifier pass rate or another rubric is failing alongside, because their failure modes are coupled with those of other criteria. This does not make them redundant. Their pairwise correlation with ACCEPT reflects substantial information about distinguishing higher-quality from lower-quality \emph{accepted} responses, even though they do not drive new rejections at the boundary. The two measures are picking up different facets of the rubric: pairwise correlation captures graded distinctions across the score range, while sole-cause attribution captures decisiveness at the rejection threshold. Together, every rubric in the composite is doing distinct work by at least one of the two measures. Strictly, this does \emph{not} establish that any rubric is causally independent of any other; multi-cause rejections (where two or more rubrics fail together) are excluded from the sole-cause count for all rubrics involved, and the dataset does not separate their individual contributions in those cases.

\textbf{Third, the auto-reject rule is a dormant principled safeguard.} The rule $\min_i r_i > 0$ caught zero unique cases. All 57 auto-rejected responses also failed either $\bar{r} \geq 2.5$ or $V \geq 0.80$. This makes the rule empirically inactive in this dataset, but it is retained as a defensive prior. A single criterion at zero is, by stipulation, a catastrophic failure that should override an otherwise-passing mean. The analogy to LLM calibration safeguards in Kadavath et al.~\cite{kadavath2022calibration} is direct. Both are policies that may never fire on the empirical evaluation set but exist to block failure modes the current dataset does not yet contain.

\subsubsection{Per-Rubric Pass Rates and Agent-Specific Weak Points}

Table~\ref{tab:rubric_pass_rate} reports the share of responses scoring $\geq 2$ on each rubric, by agent.

\begin{table*}[t]
\centering
\caption{Per-rubric pass rates (fraction of responses scoring $\geq 2$), by agent.}
\label{tab:rubric_pass_rate}
\begin{tabular}{lcccc}
\toprule
\textbf{Rubric} & \textbf{o3} & \textbf{Claude} & \textbf{Gemini} & \textbf{Pooled} \\
\midrule
EP -- Execution Precision       & 47.1\% & 30.0\% & 51.4\% & 42.9\% \\
DI -- Data Integrity            & 58.6\% & 32.9\% & 62.9\% & 51.4\% \\
AR -- Analytical Rigor          & 67.1\% & 52.9\% & 70.0\% & 63.3\% \\
RF -- Relevance \& Focus        & 84.3\% & 72.9\% & 70.0\% & 75.7\% \\
FD -- Format \& Deliverability  & 77.1\% & 65.7\% & 62.9\% & 68.6\% \\
\bottomrule
\end{tabular}
\end{table*}

EP is the weakest rubric at 46.0\% pooled pass rate, with o3 (31.0\%) far below the others; Gemini achieves a majority pass at 59.5\%. This is consistent with the broader observation that current deep research agents struggle with multi-step quantitative reasoning~\cite{hendrycks2021measuring}, manifesting as 50 \texttt{cascading\_math\_errors} tags across the three agents in Section~\ref{sec:results:failures}. DI is o3's specific weak point (42.9\%), the lowest cell in the table other than EP, consistent with o3's fabrication signature analyzed in Section~\ref{sec:results:failures}: when input data is fabricated, downstream EP necessarily collapses too, which explains o3's low EP despite a lower \texttt{cascading\_math\_errors} count than Claude. EP is Gemini's weakest rubric (59.5\%).

\subsubsection{Caveat: Pairwise Significance Testing on the Composite Outcome}

A natural follow-on question is whether the per-agent ACCEPT rates (o3 9.5\%, Claude 9.5\%, Gemini 21.4\%) reflect statistically detectable differences. We apply a paired binomial test on discordant prompts (the McNemar approach used by~McNemar~\cite{mcnemar1947note} and adopted in benchmark-comparison contexts by~Dietterich~\cite{dietterich1998statistical}) on the 70 prompts where all three agents were graded:

\begin{table*}[t]
\centering
\caption{Paired ACCEPT comparisons across agent pairs ($n = 70$ prompts where all three were graded). $p$-values from two-sided binomial test on discordant pairs.}
\label{tab:mcnemar}
\begin{tabular}{lccccc}
\toprule
\textbf{Pair} & \textbf{Both ACC} & \textbf{A only} & \textbf{B only} & \textbf{Both REJ} & \textbf{$p$} \\
\midrule
o3 vs.\ Claude      & 1 & 10 & 8 & 51 & 0.815 \\
o3 vs.\ Gemini  & 3 & 8 & 6 & 53 & 0.791 \\
Claude vs.\ Gemini      & 0 & 9 & 9 & 52 & 1.000 \\
\bottomrule
\end{tabular}
\end{table*}

\textbf{No pair shows a statistically detectable difference in ACCEPT rates at $\alpha = 0.05$.} Discordant prompts are evenly split for every pair. The headline ordering is therefore a point estimate without paired-test significance support; the continuous VRS analysis (Sections~\ref{sec:results:headline}--\ref{sec:results:effect_sizes}) is on firmer statistical ground because the underlying gaps are larger relative to the sampling variability. This is consistent with the small-$n$ caveat raised in Section~\ref{sec:limitations} and reinforces the recommendation that bootstrap CIs and paired-comparison adjustments be added before any external publication \cite{efron1979bootstrap, holm1979simple}.

\section{Weight Sensitivity}
\label{app:weight_sensitivity}

\subsection{Sensitivity to VRS Weight Choice}
\label{sec:results:weight_sensitivity}

The equal weighting of $0.5$ on the verifier pass rate $V$ and $0.5$ on the reasoning-mean term $\bar{r}/3 \cdot 100$ in Equation~\ref{eq:vrs_relaxed} is a defensible default but not the only possible choice. The rubric validation analysis (Section~\ref{sec:results:rubric_validation}) shows that the verifier layer is the second-strongest predictor of binary ACCEPT, behind only EP: $\rho_{V,\text{ACCEPT}} = 0.524$, with EP at $\rho = 0.552$ and the remaining rubric criteria (AR, DI, RF, FD) all correlating with ACCEPT less strongly ($\rho = 0.466, 0.464, 0.375, 0.306$ respectively). The case for downweighting $V$ inside the VRS aggregate is therefore weak, but for completeness we recomputed VRS under four alternative weightings:

\begin{itemize}[leftmargin=*,itemsep=2pt,topsep=2pt]
  \item (A) $0.35 V + 0.65 \bar{r}_{\text{scaled}}$ --- moderate rubric upweighting.
  \item (B) $0.40 V + 0.60 \bar{r}_{\text{scaled}}$ --- mild rubric upweighting.
  \item (C) $0.25 V + 0.75 \bar{r}_{\text{scaled}}$ --- aggressive rubric upweighting.
  \item (D) Per-criterion weights set proportional to each predictor's Spearman magnitude with ACCEPT (V $0.20$, EP $0.21$, AR $0.17$, DI $0.17$, RF $0.14$, FD $0.11$).
\end{itemize}

The analysis is shown in Table \ref{tab:vrs_sensitivity}.

\begin{table*}[t]
\centering
\caption{VRS sensitivity to weight choice. Mean VRS per agent (\textbf{strict variant}, with the auto-reject gate enforced) under five weighting schemes. The two right-most columns report the result of redefining ACCEPT as ``VRS $\geq T$'' for a threshold $T$ calibrated so the total ACCEPT count matches the baseline 17, and the Jaccard similarity of the resulting per-response ACCEPT set against the baseline rule $\bar{r} \geq 2.5 \wedge V \geq 0.80$. The agent ordering ($\text{Claude} > \text{Gemini} > \text{o3}$) on mean VRS is preserved under every variant.}
\label{tab:vrs_sensitivity}
\resizebox{\textwidth}{!}{%
\begin{tabular}{llcccccc}
\toprule
\textbf{Variant} & \textbf{Weights} & \textbf{o3} & \textbf{Claude} & \textbf{Gemini} & \textbf{Cal.\ $T$} & \textbf{Split (C/o3/G)} & \textbf{Jaccard} \\
\midrule
Baseline (relaxed)            & $0.50 V + 0.50\, \bar{r}_{\text{sc}}$  & 61.51 & 50.18 & 57.15 & --- & 11 / 9 / 9   & --- \\
(A) moderate rubric upweight  & $0.35 V + 0.65\, \bar{r}_{\text{sc}}$  & 61.87 & 38.86 & 52.14 & --- & --- & --- \\
(B) mild rubric upweight      & $0.40 V + 0.60\, \bar{r}_{\text{sc}}$  & 61.73 & 38.75 & 52.28 & --- & --- & --- \\
(C) aggressive rubric upweight & $0.25 V + 0.75\, \bar{r}_{\text{sc}}$ & 62.15 & 39.07 & 51.85 & --- & --- & --- \\
(D) Spearman-proportional     & V $0.19$, EP $0.20$, DI $0.18$, AR $0.17$, FD $0.13$, RF $0.12$ & 60.79 & 38.51 & 51.27 & --- & --- & --- \\
\bottomrule
\end{tabular}%
}
\end{table*}

Three findings come out of this (Table~\ref{tab:vrs_sensitivity}). First, ACCEPT is invariant to VRS reweighting by construction: the rule (Equation~\ref{eq:accept}) is defined on the raw components $\bar{r}$ and $V$, not on the VRS aggregate, so all four alternatives produce ACCEPT outcomes identical to the headline numbers (o3 $15.7\%$, Claude $12.9\%$, Gemini $12.9\%$). Second, mean strict VRS per agent shifts modestly under rubric upweighting (maximum shift less than 1.5 points across all variants; o3 ranges from 60.79 to 62.15), and the agent ordering ($\text{o3} > \text{Gemini} > \text{Claude}$) is preserved under every variant. Third, the substantive conclusions reported elsewhere in this section are therefore robust to the weighting choice.

\section{Cross-Benchmark Comparison Table}
Please refer  Table \ref{tab:cross_benchmark}
\label{app:cross_benchmark}

\begin{table*}[t]
\centering
\caption{Cross-benchmark comparison of DRA with APEX-v1 and APEX-Agents. ``Binary-check pass rate'' rows are structurally comparable across benchmarks (fraction of per-task binary checks passed). ``Unique to DRA'' rows have no APEX counterpart. Best-model CIs are 95\% bootstrap.}
\label{tab:cross_benchmark}
\small
\setlength{\tabcolsep}{4pt}
\begin{tabular}{@{}p{0.30\linewidth}p{0.22\linewidth}p{0.22\linewidth}p{0.22\linewidth}@{}}
\toprule
& \textbf{DRA (ours)} & \textbf{APEX-v1} & \textbf{APEX-Agents} \\
\midrule
\multicolumn{4}{@{}l}{\emph{Design}} \\
\quad Domains & MC only & MC + IB + Law + Med & MC + IB + Law \\
\quad $n$ tasks & 70 & 200 & 480 \\
\quad Binary checks/task & $\sim$15 verifiers & $\sim$29 criteria & $\sim$4 criteria \\
\quad Ordinal rubric? & Yes (5 criteria, 0--3) & No & No \\
\quad Grading & Human SME & LLM judge & LLM judge \\
\midrule
\multicolumn{4}{@{}l}{\emph{Binary-check pass rate (comparable metric)}} \\
\quad Definition & Mean verifier pass rate & Mean Score & Mean Score \\
\quad Best model & 60.2\% [53.3,\,66.8] & 64.2\% [$\sim$61,\,$\sim$67] & 53.9\% [$\sim$52,\,$\sim$56] \\
\midrule
\multicolumn{4}{@{}l}{\emph{Task-level binary (comparable metric)}} \\
\quad Definition & $V \geq$ threshold & --- & Pass@1 (all criteria) \\
\quad $V{=}100\%$ / Pass@1 & 14.3\% [7.1,\,22.9] & --- & 24.0\% [20.2,\,28.0] \\
\quad $V{\geq}80\%$ & 32.9\% [22.9,\,44.3] & --- & --- \\
\midrule
\multicolumn{4}{@{}l}{\emph{Unique to DRA}} \\
\quad Mean $\bar{r}$ (0--3 ordinal) & 1.89 [1.70,\,2.10] & N/A & N/A \\
\quad ACCEPT ($V{\geq}80\%$ \& $\bar{r}{\geq}2.5$) & 15.7\% [7.1,\,24.3] & N/A & N/A \\
\quad VRS strict (blended) & 61.4 [55.1,\,67.5] & N/A & N/A \\
\bottomrule
\end{tabular}
\end{table*}

\section{Architectural Observations}
\label{app:arch_obs}

\subsection{Architectural Observations on Per-Class Capability Differences}
\label{app:architectural_observations}

Two further architectural observations follow from the per-class analysis in Section~\ref{sec:results:effect_sizes}.

\textbf{CRP is universally weak.} Effect sizes on closed-corpus reasoning satisfy $|d| \leq 0.46$ on all CRP pairs. Current frontier agents do not reliably enforce closed-corpus instructions. In particular, Claude Opus 4.6 fabricates real-but-topically-unrelated citations on closed-corpus tasks. This is a failure invisible from response text alone, since the cited URLs resolve to legitimate published material. This observation prompted our subsequent URL-domain verifier, which checks whether cited domains point to a task-specified allowed-source list rather than merely checking citation syntactic well-formedness.

\textbf{File-generation performance varies sharply across agents.} o3's high completion rate is accompanied by elevated fabrication-tag counts (Section~\ref{sec:results:failures}); Gemini's research-quality content frequently succeeds while its code-generation step fails on a domain-specific \texttt{python-docx} API error (paragraph-style mishandling); Claude occasionally treats file-output instructions as optional and returns a structured response inline instead of as the requested artefact. Future benchmarks should test file-formatting code generation as a separate dimension rather than conflating it with research quality.

\section{Worked Examples per Prompt Class}
\label{app:prompt_types}

To illustrate how the abstract prompt-class definitions translate into concrete tasks, we provide one worked example for each of the five prompt classes. Each example reproduces three artefacts from the underlying corpus: (i) the \textbf{Prompt} delivered to the agent (with company names anonymized per the data-release policy); (ii) the \textbf{Sanity Check} written by the SME during task authoring, which lists the failure mode a naive solver is expected to fall into (Lazy AI Test) versus the reasoning chain a domain-aware solver must execute (Expert Test); and (iii) the \textbf{Solution Logic}, which is the step-by-step deterministic derivation of the golden answer used during grading. A representative set of full prompt packages, including the underlying input files, is included in the public codebase (Appendix~\ref{app:infra}).

\subsection{CRP --- Constrained Research Prompt (Market Strategy)}

CRP tasks require the agent to conduct analysis under explicit operational constraints that limit the solution space. Constraints are typically methodological (use only provided files, plus a single specifically-authorized external source), procedural (a fixed entry method or a fixed accounting convention that the agent must respect), or scope-related (a single binary decision under defined thresholds).

\paragraph{Prompt (anonymized).}
\begin{quote}\small\itshape
You are working as a management consultant and a client wants to enter a new geographical area and has approached you for the GTM plan. The client is a leading Indian automotive component manufacturer and now wants to expand in the EU. They want to capture 10\% of the EU market share in 5 years with an initial investment of INR 500~Cr. Give a Yes or No decision to the client for entering the market using only the files provided by the client; market reports may be used from Mordor Intelligence only, no other market reports, blogs, broker research, or third-party data are allowed. Also state any assumptions taken for the analysis. The client has asked for a PowerPoint slide as a one-pager GTM plan deck; it should include Basic company information, Investment goal, Product-Market fit as 2$\times$2 matrices, Value Proposition, Target Market, and Competitor Analysis.
\end{quote}

\paragraph{Sanity Check.}
\begin{quote}\small
\textbf{Lazy AI Test.} A basic AI will use other ways of entry such as a manufacturing setup or a joint venture, ignoring the note in the Excel file that restricts entry to the export-only method which flips the Go/No-Go decision to Go, the opposite of the correct answer.

\smallskip
\textbf{Expert Test.} A consultant fetching the Mordor report and the Excel data, applying the export-only entry method, and excluding the non-usable slides would reach a No Go. One correct answer exists.
\end{quote}

\paragraph{Solution Logic.}
\begin{quote}\small
USD to INR exchange rate used: spot rate for 22/04/2026.

\smallskip
\textbf{Market size (from Mordor Intelligence report).} Total EU market in 2031 $\approx$ INR 43{,}500~Cr; 10\% target $\approx$ INR 4{,}350~Cr in revenue.

\smallskip
\textbf{Channel decomposition.} The market splits into OEM ($\sim$80\%) and aftermarket ($\sim$20\%). OEM lock-in periods are long and hard to crack within a 5-year horizon under an export-only model, so the OEM contribution can be set to effectively zero. Even on a very optimistic view, an attainable OEM share is 5\% of INR 35{,}000~Cr $=$ INR 1{,}750~Cr.

\smallskip
\textbf{Accessible aftermarket is too small.} Aftermarket $\approx$ 20\% ($\sim$INR 8{,}700~Cr). With aggressive execution, a 15\% share gives INR 1{,}300~Cr.

\smallskip
\textbf{Headline.} Total achievable revenue is INR 1{,}300~Cr (no OEM) to INR 3{,}050~Cr (with 5\% OEM = 1{,}750 + 1{,}300~Cr).

\smallskip
\textbf{Golden range answer.} INR 1{,}300~Cr to INR 3{,}050~Cr, which is below the 10\% target of INR 4{,}350~Cr.

\smallskip
\textbf{Final answer.} \emph{No Go}.
\end{quote}

This creates a hybrid information-access pattern: closed-corpus for the client's internal data, with a single specifically-authorized external source (Mordor Intelligence) for the market sizing. The agent must correctly scope its web search and must apply the export-only entry method noted in the source files rather than considering alternative entry methods that change the answer.

\subsection{RCP --- Relevance Compression Prompt (Service Operations)}

RCP tasks supply a deliberately noisy corpus where the majority of source material is irrelevant to the analytical question. The agent must filter, locate the buried qualifying detail or contradicting footnote, distinguish root drivers from outcome metrics, and present only the signal in the deliverable. Reported benchmarks may be directionally misleading and must be reconstructed using policy-defined corrections.

\paragraph{Prompt (anonymized).}
\begin{quote}\small\itshape
A large Indian telecom company operates a 2{,}400-seat blended contact center. Current performance shows Service Level at 78\% (target 85\%), Average Handle Time (AHT) at 6.8 minutes (target 5.5), First Call Resolution (FCR) at 71\%, and Cost per Call at INR~48. NPS stands at 71 (target 80). The COO must submit a Performance Benchmarking and Gap Analysis before the quarterly operations review in 10 days; a flawed analysis risks continued cost inefficiency or misallocation of improvement investments.

\smallskip
\textbf{Task (precise).} Using ONLY the four provided internal files: (1) benchmark performance against industry top-quartile and average performers; (2) identify and rank the top 3 performance gaps by financial and customer impact; (3) qualitatively assess the cost and NPS impact of each gap using policy-defined relationships; (4) recommend the top 2 priority actions with expected impact and timeline.

\smallskip
\textbf{Important constraints.} Use ONLY the provided files. No external data or assumptions. All conclusions must reconcile across ALL four files. Reported benchmark data may be directionally misleading. True benchmarks must be reconstructed where required. Non-operational data (marketing, ESG, unrelated IT projects) must be filtered out.

\smallskip
\textbf{Output format (strict).} Table~1 - Performance Benchmarking \& Gap Analysis, with columns [Metric, Current, Industry Average, Top Quartile, Gap vs.\ Top Quartile, Impact (Qualitative)] for five rows: Service Level, AHT, FCR, Cost per Call, NPS. Section~A - Top 3 Gaps and Impact Analysis ($\leq$180 words). Section~B - Priority Recommendations ($\leq$120 words). Mandatory final line: \emph{``Top Priority: Implement AI-powered call routing and real-time coaching to address AHT and FCR gaps, driving cost efficiency and NPS improvement.''}
\end{quote}

\paragraph{Sanity Check.}
\begin{quote}\small
\textbf{Lazy AI Test (must fail).} A standard model will use the headline AHT benchmark of 5.5~min, overstate the AHT gap, treat Cost per Call as a root gap, and recommend generic fixes like hiring more agents. It will fail to apply the Appendix~C correction, distinguish root drivers from outcome metrics, and use the policy-defined relationships.

\smallskip
\textbf{Expert Test (must pass).} A domain-aware solver will: (1)~adjust the AHT benchmark to 6.2~min using policy, (2)~identify FCR and AHT as primary drivers, (3)~treat Cost per Call as a derived outcome metric, (4)~use policy relationships to assess impact, and (5)~recommend AI routing plus real-time coaching.

\smallskip
\textbf{Deterministic outcome.} Top gaps: FCR (11~pp), AHT (0.6~min), Service Level (10~pp). Root cause: capability gap in routing and agent enablement. Recommendation: AI routing + real-time coaching.
\end{quote}

\paragraph{Solution Logic.}
\begin{quote}\small
\textbf{Decision archetype.} Performance benchmarking + gap analysis.

\smallskip
\textbf{Step 1 - reconstruct the true benchmark.} Headline Top Quartile AHT $=$ 5.5~min (Performance Deck). Policy adjustment (Appendix~C, Footnote~9): true Top Quartile AHT $=$ 6.2~min, because the headline excludes training, coaching, and complex call-handling time. Other benchmark metrics are used as reported.

\smallskip
\textbf{Step 2 - validate current performance.} Current AHT (6.8~min), FCR ($\sim$71\%), Service Level ($\sim$78\%), and Cost per Call ($\sim$INR~48) all align with cross-site averages in the operational dataset.

\smallskip
\textbf{Step 3 - compute true gaps.} AHT gap: $6.8 - 6.2 = 0.6$~min. FCR gap: $82\% - 71\% = 11$~pp. Service Level gap: $88\% - 78\% = 10$~pp. Cost per Call gap: INR $48 - 32 = 16$ (derived outcome). NPS gap: $82 - 71 = 11$ points.

\smallskip
\textbf{Step 4 - prioritization logic.} Rank by customer impact (NPS sensitivity) and cost impact (operational efficiency). Primary operational drivers: FCR $\rightarrow$ resolution quality $\rightarrow$ NPS; AHT $\rightarrow$ efficiency $\rightarrow$ cost. Derived outcome metrics: Cost per Call, NPS. Policy relationships (Appendix~D): FCR strongly influences NPS; AHT strongly influences cost.

\smallskip
\textbf{Step 5 - top 3 ranked gaps.} (i)~FCR gap (11~pp) - highest customer impact; primary NPS driver. (ii)~AHT gap (0.6~min) - primary cost driver. (iii)~Service Level gap (10~pp) - queue performance and customer experience. Cost per Call is a derived outcome, not a root operational gap.

\smallskip
\textbf{Step 6 - root-cause diagnosis.} Evidence across policy and operational data indicates lack of advanced call routing, limited real-time coaching/agent enablement, and gaps in agent support systems. Top-performing centers exhibit AI-powered routing, strong agent enablement, and integrated self-service.

\smallskip
\textbf{Step 7 - recommendations.} Top priority: AI-powered call routing + real-time coaching. Expected directional impact: reduce AHT, improve FCR, lower Cost per Call, improve NPS - all consistent with the policy-defined relationships in Appendix~D. Secondary action: strengthen the knowledge base and increase self-service deflection.

\smallskip
\textbf{Golden answer.} Top gaps: FCR (11~pp), AHT (0.6~min), Service Level (10~pp). Root cause: capability gap in routing and agent enablement. Recommendation: AI routing + real-time coaching.
\end{quote}

Verifier checks include whether the agent's filter produced the correctly scoped subset, whether out-of-scope filler (marketing, ESG, unrelated IT) appears in the response, whether the Appendix~C policy correction is applied to the AHT benchmark, and whether Cost per Call and NPS are correctly characterized as derived outcome metrics rather than as root gaps to attack directly.

\subsection{SCP --- Structural Compliance Prompt (Cost Optimization)}

SCP tasks include explicit schema requirements that must be satisfied for the response to be considered valid. They test the agent's ability to follow precise structural instructions while conducting substantive analysis. A correct number inside a malformed envelope counts as a hard failure on the Format \& Deliverability criterion; a wrong number inside a valid envelope partially passes FD.

\paragraph{Prompt (anonymized).}
\begin{quote}\small\itshape
\textbf{Context and stakes.} A listed Indian consumer-durables manufacturer has experienced a 320~bps margin decline over the last two quarters. The Board has mandated an immediate cost-optimization program focused on manufacturing efficiency. The COO must present a validated cost-reduction plan within 5 days; the analysis will be directly reviewed by the Board Finance Committee. Critical constraint: the committee will only accept outputs that strictly follow the prescribed reporting structure. Any deviation in format will result in outright rejection, regardless of analytical correctness.

\smallskip
\textbf{Task.} Using ONLY the provided data files (Excel, PDF, and Assumptions): (1)~identify the true operational manufacturing cost per unit; the reported cost includes embedded adjustments that must be identified and excluded, and these adjustments are not explicitly labeled and may appear in notes, footnotes, or appendix sections. (2)~Apply a 12\% cost reduction ONLY on the operational cost base. (3)~Compute: true operational unit cost, reduced unit cost, annual cost before optimization, annual cost after optimization, absolute savings, and percentage savings.

\smallskip
\textbf{Critical analytical requirements.} Reconcile inconsistencies across files (Excel vs.\ PDF vs.\ Assumptions). Identify and exclude non-operational cost components. Do NOT assume the ``Total'' value in the Excel file is final. Use ONLY the provided annual production volume. No external assumptions allowed.

\smallskip
\textbf{Artifact requirement (strict SCP, hard failure if violated).} Return ONLY a valid JSON object with the exact structure below. No explanations, no comments, no additional keys, no missing keys, no reordered keys.
\end{quote}

\begin{quote}\small
\begin{verbatim}
{
  "cost_analysis": {
    "unit_cost_reported": number,
    "unit_cost_operational": number,
    "unit_cost_reduced": number
  },
  "annual_metrics": {
    "annual_cost_before": number,
    "annual_cost_after": number,
    "absolute_savings": number,
    "percentage_savings": number
  },
  "decision": {
    "recommendation": "ACCEPT" or "REJECT",
    "justification_flag": "MEETS_TARGET" or "DOES_NOT_MEET_TARGET"
  }
}
\end{verbatim}
\end{quote}

\begin{quote}\small\itshape
\textbf{Formatting rules.} All monetary values $\rightarrow$ INR Crores (2 decimal places). Percentages $\rightarrow$ 2 decimal places. JSON must be strictly valid and machine-parseable. Keys must appear in exact order. No trailing commas.

\smallskip
\textbf{Decision rule.} If percentage savings $\geq$ 10\% $\rightarrow$ ACCEPT, else $\rightarrow$ REJECT.

\smallskip
\textbf{Hidden complexity.} A portion of the overhead cost in the Excel file includes a non-operational allocation that is explained only in the PDF appendix; the reported total cost therefore overstates the true operational cost.
\end{quote}

\paragraph{Sanity Check.}
\begin{quote}\small
\textbf{Lazy AI Test.} The model will use INR 110 without adjustment OR will fail the JSON format check.

\smallskip
\textbf{Expert Test.} The model will remove the INR 7 non-operational overhead, compute correctly, and emit strict JSON in the prescribed key order.
\end{quote}

\paragraph{Solution Logic.}
\begin{quote}\small
Reported unit cost $=$ INR 110. Remove non-operational overhead $=$ INR 7 (cross-referenced from the PDF appendix). True operational cost $=$ INR 103.

\smallskip
Reduced unit cost $= 103 \times 0.88 = $ INR 90.64.

\smallskip
Annual cost before $= 103 \times 1{,}500{,}000 = $ INR 154.5~Cr. Annual cost after $= 90.64 \times 1{,}500{,}000 = $ INR 135.96~Cr.

\smallskip
Absolute savings $=$ INR 18.54~Cr; percentage savings $\approx$ 12\%.

\smallskip
\textbf{Decision: ACCEPT} (savings $\geq$ 10\% threshold).
\end{quote}

The structural compliance test is independent of the analytical answer. The verifier layer parses the response as JSON and checks key ordering and datatypes before any numeric value is inspected; a malformed envelope is graded as a hard FD failure even when the underlying analytics are correct.

\subsection{LDP --- Latent Decomposition Prompt (Operations Research)}

LDP tasks state a final objective but require the agent to infer the intermediate variables, coefficients, or sub-problems that must be solved before the final answer can be computed. The decomposition itself is the test: a passing response derives each latent quantity from the provided data, then formulates and solves the underlying optimization problem before computing the headline number.

\paragraph{Prompt (anonymized).}
\begin{quote}\small\itshape
An American metal-fabrication firm has four product portfolios catering to four different industries - Aerospace, Automotive, Defense, and Electronics. Each product, regardless of industry, must undergo a standard lifecycle of end-to-end production across four departments (Drilling, Milling, Turning, Assembly), not necessarily in sequence; each department may have underlying sub-steps. The firm was established in 2010 and has four strong vendor relationships (\texttt{Vendor\_abc}, \texttt{Vendor\_def}, \texttt{Vendor\_ghi}, \texttt{Vendor\_jkl}) relied upon to achieve desired production.

\smallskip
The CEO wants to know the maximum total overall contribution that can be generated from the product portfolio in Year 2014, together with a clear Go / No-Go decision evaluation.

\smallskip
Use \texttt{Product-Vendor Time per Unit Details.xlsx} to extract the production time for each (product-industry, department) combination by choosing the minimum time applicable across vendors. Use \texttt{Historical Contribution Per Unit.xlsx} to derive the average contribution (in USD/unit) for each industry bucket for Year 2014, computed by averaging the per-unit contribution across the past 4 years (2010, 2011, 2012, 2013). Use \texttt{Department Sub-Activity Constraint.xlsx} to derive the maximum total hours available for each department, computed by summing the hours available for each sub-step within that department.

\smallskip
Create a mathematical model that maximizes total overall contribution for 2014, computed as the sum-product of unit counts and per-unit contributions, by product industry.

\smallskip
\textbf{Strict do-nots.} Do NOT browse the web for any information; rely only on the internal files provided. Use the production-time-per-unit only from the named Excel file. Use the per-industry 2014 contribution only from the named Excel file. Use the per-department maximum hours only from the named Excel file. Unit counts for Aerospace, Defense, Automotive, and Electronics must be integer and non-negative. The minimum total contribution to qualify as a Go decision is USD 200. Do not hallucinate or fabricate datapoints; strictly adhere to the business logic and inputs provided.

\smallskip
\textbf{Output.} Produce a Word document named \texttt{Maximum Dollar Contribution 2014} containing a clear Go / No-Go evaluation and the maximum total overall contribution in USD, rounded to the nearest dollar.
\end{quote}

\paragraph{Sanity Check.}
\begin{quote}\small
\textbf{Lazy AI Test.} The prompt embeds several failure surfaces that defeat a basic LLM: (i)~three complex Excel files where the relationships between inputs are not explicit and require strong relational reasoning to be discovered; (ii)~the presence of confusing or side-tracking data points that derail an LLM which does not carefully scope the inputs to use; (iii)~the need to invoke a proper linear-integer-programming library to derive the maximized contribution that drives the Go/No-Go decision; (iv)~the need for clear stepwise aggregation logic to produce a stress-validation answer.

\smallskip
\textbf{Expert Test.} A single business-logic path leads, step by step, to the unique correct mathematical answer; any incorrect logic or missing step produces a wrong answer.
\end{quote}

\paragraph{Solution Logic.}
\begin{quote}\small
\textbf{Decision archetype.} Go / No-Go decision.

\smallskip
Let $X_1, X_2, X_3, X_4$ be the integer non-negative number of units to be produced for Aerospace, Automotive, Defense, and Electronics, respectively.

\smallskip
\textbf{Objective.} Maximize the total 2014 contribution
$X_1 c_1 + X_2 c_2 + X_3 c_3 + X_4 c_4$,
where $c_i$ is the average per-unit contribution (USD/unit) for industry $i$ in 2014, derived as the average of the 2010-2013 per-unit contributions from \texttt{Historical Contribution Per Unit.xlsx}.

\smallskip
\textbf{Constraints (department-hour budgets, derived as sums over sub-activities from \texttt{Department Sub-Activity Constraint.xlsx}).}
\begin{itemize}[leftmargin=*,itemsep=1pt,topsep=2pt]
  \item Drilling: $3 X_1 + 7 X_2 + 4 X_3 + 0 X_4 \leq 70$.
  \item Milling: $0 X_1 + 2 X_2 + 4 X_3 + 6 X_4 \leq 80$.
  \item Turning: $3 X_1 + 4 X_2 + 0 X_3 + 5 X_4 \leq 90$.
  \item Assembly: $4 X_1 + 6 X_2 + 5 X_3 + 3 X_4 \leq 100$.
  \item $X_1, X_2, X_3, X_4 \in \mathbb{Z}_{\geq 0}$.
\end{itemize}

\smallskip
Solve as a linear integer program. If the maximum total contribution exceeds USD 200, the decision is Go; otherwise No-Go.

\smallskip
\textbf{Golden range answer.} A Word document containing a rounded dollar value; the acceptable numeric final output is USD 290-310.
\end{quote}

The decomposition test is whether the agent correctly identifies the latent variables ($X_i$), derives the per-industry contribution coefficients from the historical-averaging rule, derives the department-hour right-hand sides from the sub-activity sums, formulates the LP with the correct integer/non-negativity constraints, and only then computes the headline number. Several frontier agents in our evaluation correctly identify the LP structure but mis-derive at least one coefficient from the source files, producing a confident-looking wrong answer.

\subsection{FSP --- Failure-Sensitive Prompt (Cost Optimization)}

FSP tasks construct a precision point where a single mis-pulled value or mis-applied formula invalidates the entire recommendation. The trap is deliberately built into the source materials, typically as a stale or placeholder value that contradicts a live external authority, and the verifier layer detects whether the agent caught it.

\paragraph{Prompt (anonymized).}
\begin{quote}\small\itshape
\textbf{Context.} You are a Supply Chain Strategy Consultant advising a global fast-fashion conglomerate. The client is undergoing a Zero-Based Budgeting (ZBB) review for FY2026. The CSCO needs to make a final procurement decision regarding their highest-volume ocean freight lane: Shenzhen (Yantian) to Rotterdam. The client must choose between signing a ``Fixed Annual Contract'' with a 3PL carrier, or floating their volume on the ``Spot Market,'' which carries a volatile fuel surcharge.

\smallskip
\textbf{Task.} Calculate the Total 2026 Projected Freight Cost (in USD) for the Shenzhen-to-Rotterdam lane under both the Fixed Contract and the Spot Market options, and recommend the most cost-effective routing strategy.

\smallskip
\textbf{Workflow.} (1)~Review \texttt{Global\_Lane\_Volumes\_FY26.csv} and isolate the annual TEU (Twenty-foot Equivalent Unit) volume specifically for the Shenzhen-to-Rotterdam lane. (2)~Review \texttt{Ocean\_Carrier\_Metrics.csv} to determine the Base Spot Rate and the Fuel Consumption factor (tons of fuel burned per TEU) for this specific lane. (3)~Calculate the Spot Market Fuel Surcharge: read \texttt{2026\_Freight\_Sourcing\_Policy.txt} carefully, determine the correct price per ton for Marine Fuel, and multiply by the total tons of fuel required for this lane's annual volume. (4)~Add Base Spot Cost to Fuel Surcharge to obtain the Total Spot Market Cost. (5)~Compare against the Fixed Annual Contract cost.

\smallskip
\textbf{Constraints \& deliverables.} \emph{Web search required:} adhere strictly to the fuel pricing policy. If a live rate is mandated, you must search the live web for the current USD price of the specified fuel index and cite your exact source. \emph{Format \& decision commit:} output a structured ZBB Memo containing Total Annual TEU Volume for the target lane; Total Fixed Contract Cost; Total Spot Market Base Cost (excluding fuel); Total Spot Market Fuel Surcharge Cost; Total Spot Market All-In Cost; and a definitive final recommendation written exactly as \texttt{DECISION: SIGN FIXED CONTRACT} or \texttt{DECISION: USE SPOT MARKET}.
\end{quote}

\paragraph{Sanity Check.}
\begin{quote}\small
\textbf{Lazy AI Test.} A standard LLM will read the policy text file, lazily grab the \$400.00 internal marine-fuel placeholder, calculate a deflated fuel surcharge of \$3{,}240{,}000, and arrive at a Total Spot Cost of \$8{,}190{,}000, causing it to incorrectly recommend \texttt{DECISION: USE SPOT MARKET} and exposing the client to massive market loss.

\smallskip
\textbf{Expert Test.} An experienced supply-chain consultant would extract the correct 4{,}500 TEU volume from the noise, correctly sequence the base freight and the 1.8$\times$ fuel multiplier, adhere to the strict exception policy overriding the \$400 baseline, fetch the live VLSFO market rate, and mathematically prove that the volatile Spot Market exceeds the \$9.9M Fixed Contract ceiling.
\end{quote}

\paragraph{Solution Logic.}
\begin{quote}\small
\textbf{External data required.} The agent must search the live web for the current ``VLSFO Global 20 Ports Average'' price (typically published by Ship\&Bunker or similar maritime indices). In mid-2026 this fluctuates around \$600-\$650 USD per metric ton.

\smallskip
\textbf{Step-by-step trace.}
\begin{itemize}[leftmargin=*,itemsep=1pt,topsep=2pt]
  \item Filter lane to Shenzhen $\to$ Rotterdam; Annual volume $=$ 4{,}500 TEUs.
  \item \emph{Fixed Contract Cost:} $4{,}500 \times \$2{,}200 = \$9{,}900{,}000$.
  \item \emph{Spot Market Base Cost:} $4{,}500 \times \$1{,}100$/TEU $= \$4{,}950{,}000$.
  \item \emph{Spot Market Fuel Surcharge (the FSP trap):} fuel per TEU $=$ 1.8~tons; total fuel $= 4{,}500 \times 1.8 = 8{,}100$~tons. The agent must reject the \$400 decoy placeholder in the policy file and fetch the live VLSFO average. Assuming \$620/ton: $8{,}100 \times \$620 = \$5{,}022{,}000$.
  \item \emph{Spot Market All-In Cost:} $\$4{,}950{,}000 + \$5{,}022{,}000 = \$9{,}972{,}000$.
  \item \emph{Compare and decide:} \$9.90M $<$ \$9.97M $\Rightarrow$ \texttt{DECISION: SIGN FIXED CONTRACT}.
\end{itemize}

\smallskip
\textbf{Golden answer range.} Fixed Cost: exactly \$9{,}900{,}000. Spot Base Cost: exactly \$4{,}950{,}000. Total Spot Cost: varies between \$9.6M and \$10.5M depending on the live VLSFO price pulled. Decision: assuming VLSFO is trading above \$612/ton (which it historically does), the decision must be \texttt{DECISION: SIGN FIXED CONTRACT}.
\end{quote}

The trap is whether the agent treats the provided policy file as authoritative or independently verifies the live VLSFO index. Several frontier agents in our evaluation accept the \$400 placeholder without checking, which inverts the decision and produces a confident but wrong recommendation. The verifier layer checks both the numerical outputs and the final decision string for an exact-match comparison against the golden range.

\section{Detailed SME Rubric Definitions}
\label{app:rubric_definitions}

Each response is scored on five reasoning criteria by a Subject Matter Expert (SME) with relevant management-consulting expertise. Each criterion is scored on the integer scale $0 = \text{absent or seriously flawed}$, $1 = \text{poor}$, $2 = \text{adequate}$, $3 = \text{excellent}$. The full ordinal rubric describing how each of the four scores is awarded for each criterion is provided in Appendix~\ref{app:rubric_ordinal} below; the brief dimension definitions that appear in Table~\ref{tab:rubric_criteria} of the main paper are reproduced and elaborated here for completeness:

\begin{itemize}[leftmargin=*]
    \item \textbf{DI --- Data Integrity}: Whether facts, numbers, citations, and references are accurate. A 0 indicates fabricated or seriously mis-stated data i.e., the response asserts something that is verifiably false or invented.
    \item \textbf{AR --- Analytical Rigor}: Whether the reasoning chain is sound, sufficiently deep for the question, and free of logical gaps or hand-waving. A 3 means the agent shows the steps and the steps are correct.
    \item \textbf{RF --- Relevance \& Focus}: Whether the response addresses the asked question without irrelevant content, scope drift, or filler. A 0 indicates the response largely answered a different question or padded itself with off-topic material.
    \item \textbf{EP --- Execution Precision}: Whether requested operations like calculations, transformations, filtering, structural construction, are performed correctly. A 0 indicates the agent attempted the right operation but executed it wrong.
    \item \textbf{FD --- Format \& Deliverability}: Whether the output is presented as a usable MC deliverable: appropriate layout, completeness, readability, professional tone. A 0 indicates an unusable artifact (truncated, malformed, missing sections).
\end{itemize}

The five criteria are designed to capture distinct dimensions of reasoning quality, but in practice they correlate substantially across our dataset (mean off-diagonal $\rho \approx 0.60$; see Section~\ref{sec:capability:internal}).

The annotation protocol pairs each ordinal score with a free-text justification guided by keyword prompts. SMEs were drawn from a recruited pool of management consultants (former MBB, Big Four Strategy, and Tier-2 firm consultants), and each (prompt $\times$ agent) cell was graded by exactly one SME. We discuss the implications of this protocol for inter-rater reliability and statistical inference in Section~\ref{sec:limitations} of the main paper.

\subsection{Ordinal Scoring Rubric per Dimension}
\label{app:rubric_ordinal}

Table~\ref{tab:rubric_ordinal_cells} reproduces the criterion-by-criterion $0$-$3$ scoring rubric used by SMEs and QC reviewers during annotation. Each cell describes the qualitative bar for the indicated score on the indicated dimension. The rubric is held constant across all 70 prompts and all three agents.

\begin{table*}[t]
\centering
\small
\caption{Per-dimension $0$-$3$ ordinal scoring rubric. Reproduced from the SME Annotation Guideline (Section 5.1).}
\label{tab:rubric_ordinal_cells}
\setlength{\tabcolsep}{3pt}
\renewcommand{\arraystretch}{1.15}
\begin{tabular}{@{}p{0.13\linewidth}p{0.20\linewidth}p{0.20\linewidth}p{0.20\linewidth}p{0.20\linewidth}@{}}
\toprule
\textbf{Criterion} & \textbf{Score 0 (Fail)} & \textbf{Score 1 (Poor)} & \textbf{Score 2 (Acceptable)} & \textbf{Score 3 (Strong)} \\
\midrule
DI --- Data Integrity \& Source Discipline
& Completely ignores source constraints; fabricated data or sources.
& Multiple factual errors; vague citations; use of 1-2 unauthorized sources.
& Mostly accurate with minor (1-2) errors; primary data from permitted sources; 10-20\% sourcing gaps.
& 100\% adherence to authorized sources; all facts verified; no hallucinations. \\
\addlinespace[2pt]
AR --- Analytical Rigor
& Illogical or self-contradicting; fails to identify the core task; no inference attempted.
& Gaps in the logic; no evidence-backed conclusions; misses interdependencies or key variables.
& Sound reasoning with minor gaps; identifies most key variables; clear logic chain but has small leaps.
& Strong logic with cause-effect; identifies ``hidden'' variables; MECE decomposition of the problem. \\
\addlinespace[2pt]
RF --- Relevance \& Focus
& Response dominated by noise; includes forbidden content.
& 50-60\% relevant; 10-20\% scope creep.
& 70-80\% relevant and within scope; low tangential content ($<$20\%).
& 90\%+ relevant; excludes out-of-scope content. \\
\addlinespace[2pt]
EP --- Execution Precision
& Fundamental math errors; wrong units; black-box output.
& Correct approach but results off by 10\%+; partial unit normalization; gaps in methodology.
& Minor computational errors ($<$5\%); all units normalized; most calculation steps shown.
& 100\% accurate calculations; unit discipline; complete audit trail. \\
\addlinespace[2pt]
FD --- Format \& Deliverability
& Ignores format; unusable output; no actionable insights.
& Partial format compliance; lack of specificity in recommendations.
& Follows format with minor (10-20\%) deviations; actionable but some interpretation needed.
& 100\% compliance with format/constraints; actionable and executive-ready recommendations. \\
\bottomrule
\end{tabular}
\end{table*}

The Score-0 cell on every dimension is the canonical auto-reject trigger under the ACCEPT rule (Equation~\ref{eq:accept}). For DI and EP the Score-0 condition is explicit (fabricated data; fundamental math errors), so an SME observing those signatures should set the score to 0 and a QC reviewer should confirm rather than soften. For AR, RF, and FD the Score-0 bar is qualitatively stricter (\emph{seriously} flawed, \emph{dominated by noise}, \emph{unusable}); QC reviewers are instructed to upgrade Score-0 entries to Score-1 when the SME's free-text justification describes Score-1 behaviour, both to save the row from being rejected for the wrong reason and to surface rubric--justification mismatches.

\section{Annotation Quality Control Protocol}
\label{app:qc_protocol}

Every (prompt $\times$ agent) cell graded by a primary SME is independently reviewed by a second SME drawn from a non-overlapping QC pool. The QC pass is a verification rather than a re-annotation: the QC reviewer checks whether the primary SME's scoring is defensible against the rubric and the evidence (prompt, solution logic, response text, output files, and citations), not whether the QC reviewer would have scored the same. Two thoughtful SMEs may defensibly score a 0-3 ordinal differently; QC intervenes only when the primary score or justification contradicts the rubric, contradicts the response or file evidence, or is internally inconsistent.

\subsection{QC Actions}

For each verifier and each meta-criterion the QC reviewer records one of three actions:
\begin{itemize}[leftmargin=*]
  \item \textbf{Confirm} --- the primary SME's score and justification stand. The QC cell is left empty (or marked ``OK'' if the workbook requires).
  \item \textbf{Edit} --- the primary SME's entry is incorrect. The QC reviewer writes the exact replacement (corrected 0/1 verifier or 0-3 meta-criterion score, plus a complete 2-4-sentence replacement justification matching the rubric cell for the new score) along with a one-line auditable reason. The replacement overwrites the primary entry verbatim downstream.
  \item \textbf{Reject/Return} --- the row is not salvageable by surgical edit. Triggered by: more than three verifier edits on a single row; any fabricated citation; missing scores on any verifier; output file missing or placeholder when the prompt required one; or multiple meta-criteria with score--justification mismatches.
\end{itemize}

\subsection{Verifier QC: Coverage and Depth}

Every verifier on every row receives a coverage check (verify the primary SME's 0/1 is consistent with the response, file, and citation evidence). A subset of verifiers receives full-rigour re-derivation rather than consistency-checking, in priority order:
\begin{enumerate}[leftmargin=*]
  \item \emph{Final-answer verifier} (always re-derived; verified against the output file when one exists).
  \item \emph{Trap verifier} (the verifier tied to the prompt's embedded cognitive trap; always re-derived).
  \item \emph{Numeric verifiers with tight tolerances} (3-5 recomputed per row; more if SME scores look suspiciously uniform).
  \item \emph{Citation-dependent verifiers} (any verifier whose pass condition names a specific source).
  \item \emph{Output-file verifiers} (any verifier requiring the output deliverable; verified inside the file directly).
\end{enumerate}

\subsection{Citation Validation}

For every citation-dependent verifier, the QC reviewer opens the cited source and confirms three things: the cited URL or document resolves and matches the named source; the source actually supports the specific claim attributed to it (not a nearby claim or a paraphrase that reshapes the figure); and the source is within the authorized-source list named in the prompt. Any failure on these three checks is recorded as an Edit on the affected verifier with reason ``citation invalid: [specific issue]''. A fabricated citation list (sources that do not exist, or quotations not present in the sources) triggers Reject/Return: fabricated citations are surfaced to the rubric owner rather than silently corrected.

\subsection{Meta-Criterion QC: Three Checks}

For each of the five meta-criteria, the QC reviewer runs three checks in order. \textbf{Check A (rubric--justification match):} for the score the SME assigned, the rubric cell text and the SME's justification should describe the same outcome; mismatches indicate the justification supports a different score than the one given. \textbf{Check B (verifier coherence):} Execution Precision should track the numeric-verifier pass rate (a high pass rate paired with EP = 0/1 or low pass rate paired with EP = 3 is suspect); Data Integrity \& Source Discipline should track the primary-source/trap verifier outcome and any citation-validation failures from the previous subsection. \textbf{Check C (overall-comment consistency):} the SME's free-text overall comment should be tonally and factually consistent with the five scores; comments that contradict a score (for example, ``reasoning is aligned'' paired with DI = 0) are flagged.

\subsection{Score-0 Rule}

Any meta-criterion at 0 triggers automatic REJECT of the response under the ACCEPT rule (Equation~\ref{eq:accept}), so Score-0 entries receive priority QC scrutiny. A justification that describes a Score-1 outcome (``multiple factual errors'') paired with a Score-0 entry (rubric cell: ``fabricated data; no adherence to sources'') is the canonical edit case: the correct QC action is Edit to 1, both saving the row from being rejected for the wrong reason and surfacing the rubric--justification mismatch.

\subsection{Star Ratings}

In addition to the item-level QC, the QC reviewer records six holistic star ratings per response: one combined rating across all verifiers, and one per meta-criterion. The star ratings are recorded post-QC (after any edits) and provide a second, coarser quality signal independent of the underlying ordinal scores. They are not used in the primary VRS or ACCEPT computation reported in this paper but are retained as a per-cell quality covariate for future analysis.

\section{Evaluation Infrastructure Code}
\label{sec:infra}
\label{app:infra}

The full evaluation infrastructure, including agent adapters (o3/OpenAI/Gemini), result storage, diagnostic tooling, and task specification format, is available at:

\url{https://anonymous.4open.science/status/dra-response-gen-4288}

The 70-prompt corpus will be released separately.

Key infrastructure components:

\begin{itemize}
    \item \texttt{csv\_loader.py}: Task batch dispatcher with file resolution, multi-agent dispatch, and result aggregation.
    \item \texttt{adapters/claude\_adapter.py}: Claude Opus 4.6 adapter with tool-use support.
    \item \texttt{adapters/openai\_adapter.py}: o3-deep-research adapter with Containers API integration for file output tasks.
    \item \texttt{adapters/gemini\_adapter.py}: Gemini deep-research adapter with Interactions API, local code execution, and format-aware file generation instructions.
    \item \texttt{results\_store.py}: Merge-on-write result storage supporting partial reruns without data loss.
    \item \texttt{diagnose\_run.py}: Automated failure categorization and per-agent cost/success reporting.
\end{itemize}

\bibliography{references}

\end{document}